\documentclass[10pt,twocolumn,letterpaper]{article}

\usepackage[pagenumbers]{cvpr}
\usepackage{multirow}
\usepackage{booktabs}
\usepackage{newtxtext,newtxmath}
\usepackage{bbm}
\usepackage{svg}      
\svgsetup{inkscapelatex=true} 
\usepackage[table]{xcolor}  
\usepackage{caption} 

\definecolor{cvprblue}{rgb}{0.21,0.49,0.74}
\usepackage[pagebackref,breaklinks,colorlinks,allcolors=cvprblue]{hyperref}

\def\paperID{18877} 
\def\confName{CVPR}
\def\confYear{2026}

\title{LAOF: Robust Latent Action Learning with Optical Flow Constraints}

\author{
	Xizhou Bu$^{1}$, Jiexi Lyu$^{1}$, Fulei Sun$^{1}$, Ruichen Yang$^{1}$, Zhiqiang Ma$^{2}$, Wei Li$^{1,\dagger}$ \\
	$^{1}$Fudan University, $^{2}$Northwestern Polytechnical University
}

\begin{document}
\maketitle
\begingroup
\renewcommand\thefootnote{}
\footnotetext{$^\dagger$Corresponding author}
\endgroup
\begin{abstract}
Learning latent actions from large-scale videos is crucial for the pre-training of scalable embodied foundation models, yet existing methods often struggle with action-irrelevant distractors. Although incorporating action supervision can alleviate these distractions, its effectiveness is restricted by the scarcity of available action labels. Optical flow represents pixel-level motion between consecutive frames, naturally suppressing background elements and emphasizing moving objects. Motivated by this, we propose robust \textbf{L}atent \textbf{A}ction learning with \textbf{O}ptical \textbf{F}low constraints (LAOF), a pseudo-supervised framework that leverages the agent’s optical flow as an action-driven signal to learn latent action representations robust to distractors. Experimental results show that the latent representations learned by LAOF outperform existing methods on downstream imitation learning and reinforcement learning tasks. This superior performance arises from optical flow constraints, which substantially stabilize training and improve the quality of latent representations under extremely label-scarce conditions, while remaining effective as the proportion of action labels increases to 10\%. Importantly, even without action supervision, LAOF matches or surpasses action-supervised methods trained with 1\% of action labels. Code can be found at \url{https://github.com/XizoB/LAOF}
\end{abstract}    
\section{Introduction}
\label{sec:intro}

In recent years, latent action learning based on large-scale, action-free datasets has attracted increasing attention for its superior pretraining efficiency and scalability \cite{mccarthy2025towards, yu2025survey, ai2025review}. Among existing approaches \cite{edwards2019imitating, schwarzer2020data, schmidt2023learning, ye2022become, liu2025stamo}, the Latent Action Policies (LAPO) paradigm \cite{schmidt2023learning} has played a pivotal role in enabling the unsupervised training of latent action models (LAMs), facilitating scalable embodied foundation models \cite{ye2024latent, bjorck2025gr00t, bu2025agibot, huang2025enerverse}. The core idea of LAPO is to jointly train an inverse dynamics model (IDM) and a forward dynamics model (FDM) within an autoencoding framework using a reconstruction objective, allowing latent actions to encode the changes between consecutive observations.

However, the original LAPO implicitly assumes that all changes between consecutive observations are solely caused by the agent’s actions, as in scenarios like robot manipulation on static backgrounds \cite{o2024open, khazatsky2024droid, liu2023libero}. Unfortunately, this assumption does not hold for real-world large-scale videos, where environmental dynamics are not exclusively driven by the agent but may also rise from stochastic factors, such as moving background objects \cite{stone2021distracting}. Moreover, relying solely on reconstruction without explicit physical constraints can lead latent action representations to become entangled with visual appearance, rather than accurately capturing the agent’s physical motion.

To address these, existing methods incorporate action supervision by decoding latent actions into ground-truth actions during training, thereby reducing the impact of distractors and enhancing physical consistency \cite{zhang2025latent, nikulin2025latent, liang2025clam, chen2025villa}. LAOM \cite{nikulin2025latent} and its follow-up works \cite{liang2025clam, zhang2025latent} demonstrate that even a small set of action labels can significantly improved downstream performance. Building on this, villa-X \cite{chen2025villa} additionally introduces a proprioceptive FDM that decodes latent actions into physical actions and predicts the robot’s future low-level proprioceptive states, further strengthening the coupling between latent actions and the agent’s physical motion. These methods are typically trained by alternating between labeled and unlabeled data \cite{nikulin2025latent, liang2025clam, zhang2025latent}. Nonetheless, in real-world scenarios where action labels are extremely scarce compared to readily available large-scale videos, this alternating training often becomes unstable and prone to overfitting. Under such sparse supervision, the model easily drifts toward spurious correlations in visual features \cite{nikulin2025latent}, hindering the anchoring of latent actions in a physically meaningful action space.

In this work, we propose LAOF, a pseudo-supervised framework that leverages the agent’s optical flow as an auxiliary action signal to constrain latent action learning. LAOF is founded on the key insights that (1) the agent's optical flow provides explicit pixel-level motion between consecutive frames, which is inherently correlated with ground-truth actions; and (2) advances in optical flow estimation have produced robust models with strong cross-scene generalization, enabling optical flow to serve as an effective supervision signal without requiring manual annotation \cite{wang2024sea, teed2020raft}. Specifically, LAOF employs a dedicated flow decoder to directly map latent actions into optical flow, thereby imposing a physical motion constraint. The optical flow pseudo-labels used for training are generated by a pre-trained optical flow foundation model. For large-scale datasets with limited action labels, we propose an action-supervised method, LAOF-Action, which further incorporates an action decoder to apply supervision on labeled data while using optical flow to provide complementary supervision on unlabeled data. Our main contributions are as follows:
 
(1) We introduce optical flow constraints to provide pseudo-supervision of physical motion, enabling more stable and robust latent action learning, and validate their effectiveness on downstream imitation learning tasks on LIBERO and reinforcement learning tasks on PROCGEN.

(2) We evaluate the effect of adding optical flow constraints to the baseline on datasets with gradually increasing action-label proportions, showing that they remain beneficial even at a 10$\%$ action ratio, and that LAOF matches or outperforms action-supervised methods trained with a 1$\%$ action ratio, even without action supervision.

(3) We conduct an ablation study that varies both the constraint structure and its coupling with the FDM to explore various LAOF variants, and find that directly attaching a dedicated flow decoder to the latent actions yields the best performance.

\begin{figure*}[t!]   
	\centering
	\includegraphics[width=\textwidth]{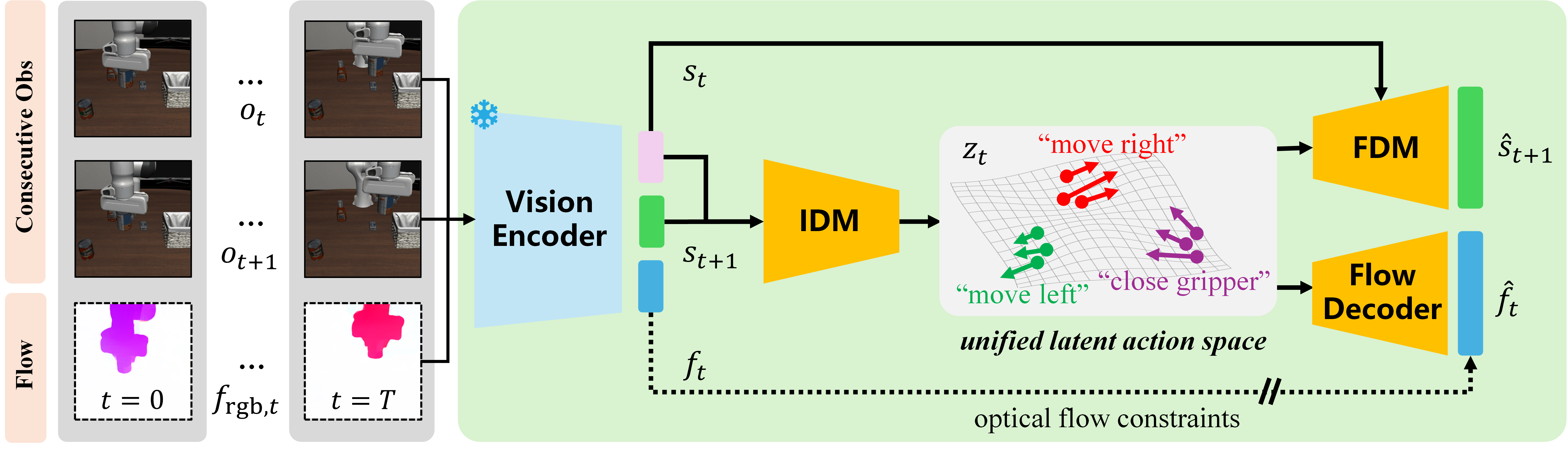}
	\caption{Overview of LAOF framework: Consecutive observations $(o_t, o_{t+1})$ and their corresponding RGB-formatted optical flow $f_{\text{rgb},t}$ are encoded into feature space $(s_t, s_{t+1}, f_t)$. The inverse and forward dynamics models, along with the flow decoder, are then jointly optimized under the combined supervision of next-state reconstruction and optical flow constraints.}
	\label{fig:laof}
\end{figure*}
\section{Related Work}
\textbf{Latent action forms.} The form of latent action representations greatly affects their quality. While continuous representations are highly expressive, they make the IDM prone to shortcut learning \cite{yang2025learning}, often degenerating into future-frame encoding rather than capturing true inter-frame changes. In response, many studies \cite{schmidt2023learning, chen2024igor, chen2024moto, cui2024dynamo, ye2024latent, bu2025agibot, bjorck2025gr00t, chen2025villa, bruce2024genie, bu2025learning} adopt the VQ-VAE \cite{van2017neural} quantization operation to discretize latent actions, creating an information bottleneck that encourages the IDM to focus on dynamics-relevant features. Yet, the relative advantages of discrete and continuous representations remain debated, with recent studies suggesting that continuous ones generally achieve better performance \cite{garrido2026learning, yang2025learning, liang2025clam, zhang2025latent, nikulin2025latent, gao2025adaworld}. For example, CoMo \cite{yang2025learning} employs a temporal difference strategy on continuous representations, replacing the IDM’s future-frame input with inter-frame differences to avoid shortcut learning. LAOM \cite{nikulin2025latent} demonstrates that continuous representations with a larger bottleneck improve both ground-truth actions predictability and downstream policy performance, a finding corroborated by our experiments. Our work provides a comprehensive evaluation of different representation forms across both imitation learning and reinforcement learning tasks.

\noindent\textbf{Visual encoder.} A growing research trend is to transform raw images into a compact feature space before training the dynamics models \cite{bruce2024genie, cui2024dynamo, nikulin2025latent, yang2025learning}. DynaMo \cite{cui2024dynamo} and LAOM \cite{nikulin2025latent} decouple the visual encoder from the IDM, with the FDM operating in feature space by reconstructing image features rather than high-dimensional pixel-level observations. Genie \cite{bruce2024genie} independently trains the visual encoder and CoMo \cite{yang2025learning} directly utilizes a pretrained ViT \cite{dosovitskiy2020image} as an image feature extractor. These works show that training IDM and FDM in feature space improves both representation quality and transferability across tasks and environments. Accordingly, we employ DINOv2 \cite{oquab2023dinov2}, a pre-trained visual encoder that provides robust and generalizable image features for downstream tasks, to control for variations in visual encoder learning across environments. Futhermore, some works \cite{klepach2025object, villar2025playslot} use pre-trained visual foundation models to decompose input video frames into spatio-temporal object slots for learning object-centric latent actions, mitigating performance degradation caused by distractors. Compared with spatio-temporal object slots, object-centric optical flow isolates objects and provides more direct motion information, allowing it to serve as a supervision signal that imposes stronger constraints on latent action learning.

\noindent\textbf{Capability extension.} 
The capability extension of LAPO have primarily evolved along two directions. The first builds on the IDM branch, leading to the development of visual-language-action models (VLAs). For example, LAPA \cite{ye2024latent} extends the original LAPO by integrating language modalities to achieve multi-task learning, while UniVLA \cite{bu2025learning} employs DINOv2 \cite{oquab2023dinov2} to construct task-centric latent action representations, mitigating the influence of task-irrelevant dynamics. GO-1 \cite{bu2025agibot} scales up the training of LAMs to millions of trajectories, and GR00T N1 \cite{bjorck2025gr00t} further enlarges the data scale by training LAMs on a heterogeneous mixture of real robot trajectories, human videos, and synthetic data. It employs a diffusion transformer module as the action head to build a large-scale embodied foundation model. The second direction builds on the FDM branch, inspiring the development of world models (WMs). For instance, Genie \cite{bruce2024genie} trains an interactive world model in the image feature space, 
whereas AdaWorld \cite{bu2025learning} develops an autoregressive world model based on diffusion modeling, and DreamDojo\cite{gao2026dreamdojo} further scales this framework by expanding the training data to 44k hours of egocentric human videos. In addition, several interesting approaches have emerged, such as JIF \cite{khandate2025train}, which integrates tactile inputs to introduce more physical priors, and UniSkill \cite{kim2025uniskill}, which reduces distillation loss by providing future-frame information via video prompts, enabling the IDM to directly infer latent actions. A contemporaneous work that also uses optical flow is FlowVLA \cite{zhong2025flowvla}, but it discretizes optical flow into tokens for autoregressive FDM training to improve world model quality and requires language supervision, whereas LAOF avoids discretization, using optical flow as the continuous constraints to learn robust latent actions without action labels or language supervision, aiming to extend the LAPO paradigm.

\section{Problem Setting}
This work aims to leverage optical flow constraints to effectively capture the priors of an agent’s motions, thereby learning robust latent actions from videos. Formally, the observation space is defined as $\mathcal{O}=\mathbb{R}^{H\times W\times 3}$, and the optical flow space as $\mathcal{F}_{\text{rgb}}=\mathbb{R}^{H\times W\times 3}$, where $H$ and $W$ denote image height and width, respectively. The raw optical flow is converted into a standard RGB format to enable unified modeling, as detailed in Section~\ref{subsec:flow2rgb}. The state space is defined as $\mathcal{S} = \mathbb{R}^{d}$, where $d$ denotes the dimensionality of the state. The latent action space is defined as $\mathcal{Z}=\mathbb{R}^{k}$, and the physical action space as $\mathcal{A}=\mathbb{R}^{n}$, where $k$ and $n$ denote the dimensions of latent and physical actions, respectively.

\section{Methodology}
The training process is structured as a three-stage pipeline: pre-training, distillation, and fine-tuning, each corresponding to a specific dataset. The unlabeled dataset, $\mathcal{D}_{\text{unlabel}}=\{(o_t, f_{\text{rgb},t})\}_{t=1}^{N}$, is used in the pre-training stage, where $o_t$ and $f_{\text{rgb},t}\in\mathcal{F}_{\text{rgb}}$ denote the observation and optical flow, respectively. It is important to note that the primary objective of pre-training is to learn high-quality latent action representations capable of capturing inter-frame transitions. Accordingly, the dataset can include partially failed demonstrations or videos without clear task relevance \cite{liang2025clam}. In the distillation stage, high-quality data containing the task’s language instruction $l_t$ is essential for transferring the learned representations to the latent action policy. We construct the latent-labeled dataset $\mathcal{D}_{\text{latent}}=\{(o_t, z_t, l_t)\}_{t=1}^{N}$ by using the frozen pre-trained IDM to annotate each consecutive observation pair $(o_t, o_{t+1})$ from this dataset with its corresponding latent action $z_t$. Finally, only  a small action-labeled dataset $\mathcal{D}_{\text{action}}=\{(o_t, a_t, l_t)\}_{t=1}^{M}$, is used for fine-tuning, where $a_t$ denotes the physical action. Here, $N$ and $M$ denote the numbers of unlabeled and labeled data, respectively. In our experiments, we consider two dataset settings: (1) an extreme setting with pure videos, and (2) a more realistic setting where $N \gg M$, i.e., large-scale videos with limited action labels.

\subsection{Pseudo Supervision from Optical Flow}
The LAOF framework integrates the inverse and forward dynamics models along with a flow decoder $d_{\text{flow}}:\mathcal{Z}\rightarrow\mathcal{F_{\text{rgb}}}$, as illustrated in Fig.\ref{fig:laof}. Here, IDM is implemented as a spatial-temporal transformer \cite{xu2020spatial}, whereas FDM and the flow decoder are implemented as spatial transformers. Based on the assumption that an agent’s optical flow is highly correlated with its physical actions, we leverage inter-frame optical flow to obtain a latent action space $\mathcal{Z}$ that approximates the physical action space $\mathcal{A}$.

\noindent\textbf{Pre-training.} An inverse dynamics model $p_{\text{IDM}}(z_t | s_t, s_{t+1})$ is trained to infer the latent action from a pair of consecutive states $(s_t, s_{t+1})$, while a forward dynamics model $p_{\text{FDM}}(\hat{s}_{t+1}|s_t, z_t)$ predicts the next state conditioned on the current state and the inferred latent action. In the original LAPO framework, these two models are jointly optimized by minimizing the next-state reconstruction loss: $\mathcal{L}_{\text{reconstruction}}(t) := \|\hat{s}_{t+1} - s_{t+1}\|_2.$ We extend this framework by introducing a flow decoder $d_{\text{flow}}(\hat{f}_t|z_t)$, which decodes the latent action into the agent’s optical flow features $\hat{f}_t\in \mathcal{S}$, thereby providing an additional motion constraint. The decoder is supervised using pseudo-labeled optical flow via the loss: $\mathcal{L}_{\text{flow}}(t) := \|\hat{f}_{t} - f_{t}\|_2$, where $f_t$ denotes the RGB-formatted optical flow $f_{\text{rgb},t}$ encoded into the state space through the visual encoder. The overall pre-training objective in LAOF combines both the next-state reconstruction loss and the optical flow constraint loss.

\begin{equation}
\mathcal{L}_{\mathrm{pretrain}} = \mathcal{L}_{\mathrm{reconstruction}} + \mathcal{L}_{\mathrm{flow}}.
\end{equation}

\begin{figure*}[t!]   
	\centering
	\includegraphics[width=\textwidth]{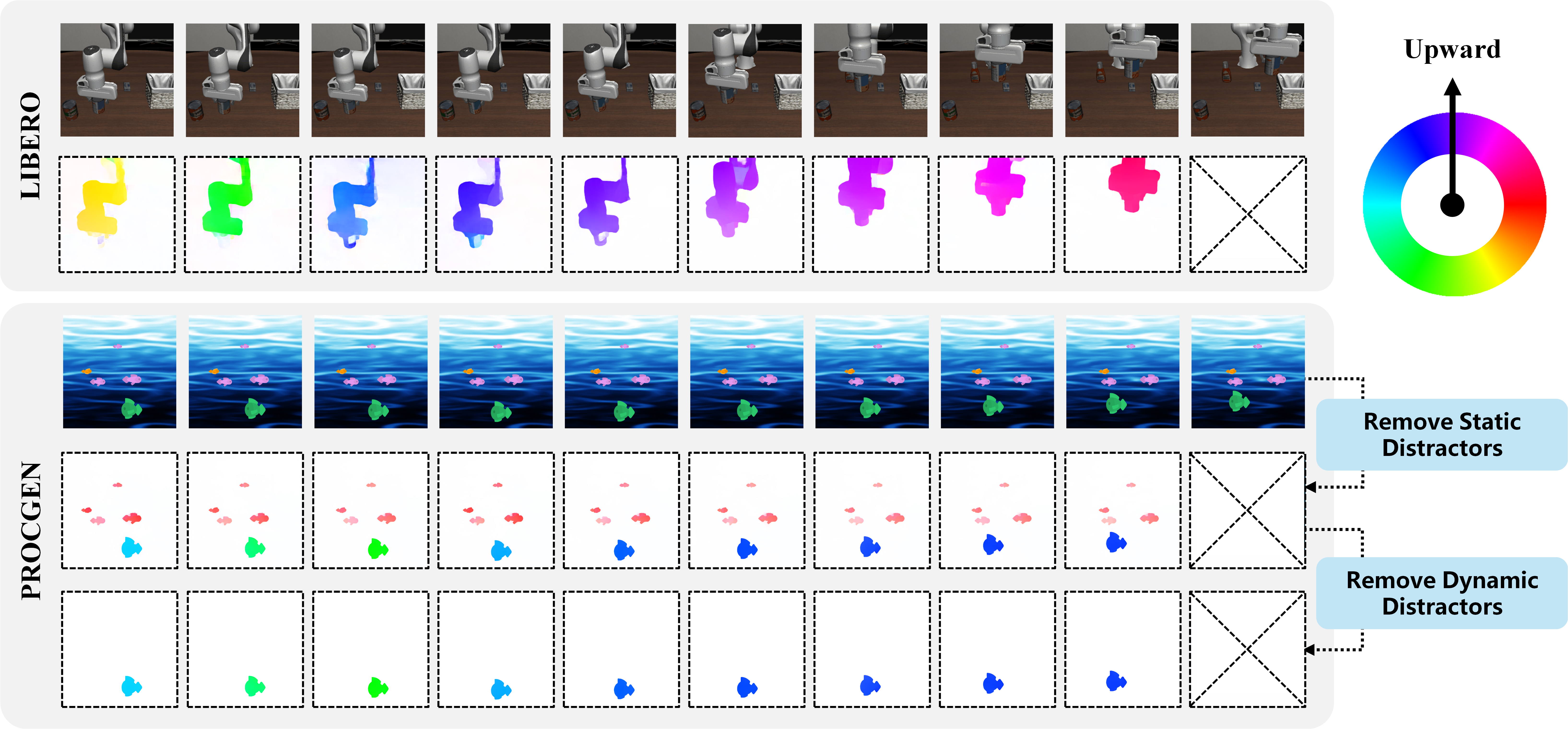}
	\caption{Visualization of optical flow on LIBERO and PROCGEN. Inter-frame optical flow, estimated using RAFT \cite{teed2020raft}, is shown below each image, representing the motion from the current frame to the next. Colors indicate motion direction, with purple corresponding to upward movement. In tasks where all distractors are static, the agent’s optical flow can be directly extracted, as illustrated by the robotic arm motions in LIBERO. For scenarios involving dynamic distractors, LangSAM \cite{langSAM} is employed to isolate object-centric optical flow.}
	\label{fig:of}
\end{figure*}

\noindent\textbf{Distillation.} This stage enables the latent policy to infer latent actions directly from the current state, effectively transferring the learned IDM representations into a deployable policy without relying on future frames. To this end, we initialize $\pi$ and perform behavior cloning using the latent-labeled dataset $\mathcal{D}_{\text{latent}}$, which is then optimized via gradient descent on the following distillation loss: $\mathcal{L}_{\mathrm{distillation}} := \|\pi(\hat{z}_t|s_t,l_t) - z_t\|_2,$ where $z_t \sim p_{\text{IDM}}(\cdot|s_t, s_{t+1})$.

\noindent\textbf{Fine-tuning.} To deploy the latent action policy on a real agent, we use a small action-labeled dataset $\mathcal{D}_{\text{action}}$ to train an action decoder $d_{\text{action}}:\mathcal{Z}\rightarrow\mathcal{A}$, which maps latent actions to physical actions. The decoder is optimized using the following loss: $\mathcal{L}_{\text{action}} := ||d_{\text{action}}(\hat{a}_t|z_t)-a_t||_2$, where $z_t \sim \pi(\cdot \mid s_t, l_t)$ is kept frozen. After training, the decoder and the latent action policy can be composed into a single function, $d_{\text{action}} \circ \pi$, which provides an end-to-end mapping from states to physical actions conditioned on the given language instruction.

\subsection{Learning with Sparse Action Supervision}
We introduce LAOF-Action for scenarios where a small set of action labels is available in the training dataset. Building upon the base LAOF framework, it incorporates an action decoder that applies explicit action supervision to the labeled samples, thereby enhancing the consistency between latent actions and physical actions. For the remaining unlabeled data, LAOF-Action provides complementary supervision through optical flow, which serves as an auxiliary action signal to constrain latent action learning. The action decoder is implemented as a lightweight multilayer perceptron (MLP). Accordingly, the overall pre-training objective of LAOF-Action is formulated as:
\begin{equation}
	\mathcal{L}_{\mathrm{pretrain}} =  \mathcal{L}_{\mathrm{reconstruction}}  + (1-\lambda)\cdot \mathcal{L}_{\mathrm{flow}}  + \lambda\cdot \mathcal{L}_{\mathrm{action}},
\end{equation}
where $\lambda$ balances the contributions of action supervision and optical flow constraints. Following \cite{zhang2025latent}, we set $\lambda = \frac{M}{N+M}$, which corresponds to the proportion of action labels in the overall training dataset. The training process follows an alternating update strategy: in each iteration, the unlabeled data are used to update the flow decoder, while the action-labeled data are used to update the action decoder.

\begin{figure*}[t!]
	\centering
	
	\begin{minipage}{\linewidth}
		\centering
		\captionof{table}{Effect of continuous latent actions on downstream imitation learning performance on LIBERO. MSE denotes the mean squared error between the predicted and ground-truth actions, Succ. denotes the average task success rate over 1000 trials, w/ OF denotes that the method uses optical flow constraints, and Avg. Impr. indicates the average improvement over LAPO. LAOM-Action and LAOF-Action are action-supervised methods, evaluated under a 1$\%$ action ratio.}
		\label{tab:libero_action_classification}
		\small 
		\setlength{\tabcolsep}{3pt}       
		\renewcommand{\arraystretch}{0.9} 
		\begin{tabular}{lcccccccccccc}
			\toprule
			\multirow{2}{*}{Method} 
			& \multicolumn{2}{c}{SPATIAL} 
			& \multicolumn{2}{c}{OBJECT} 
			& \multicolumn{2}{c}{GOAL} 
			& \multicolumn{2}{c}{LONG} 
			& \multicolumn{2}{c}{Avg. Impr.} \\
			\cmidrule(lr){2-3} \cmidrule(lr){4-5} \cmidrule(lr){6-7} \cmidrule(lr){8-9} \cmidrule(lr){10-11}
			& MSE ($\downarrow$) & Succ. ($\uparrow$) & MSE ($\downarrow$) & Succ. ($\uparrow$) & MSE ($\downarrow$) & Succ. ($\uparrow$) & MSE ($\downarrow$) & Succ. ($\uparrow$) & MSE ($\downarrow$) & Succ. ($\uparrow$) \\
			\midrule
			LAPO \cite{schmidt2023learning} 
			& $0.162$ & $80.4 \pm 1.7$
			& $0.139$ & $81.2 \pm 2.4$
			& $0.219$ & $84.0 \pm 2.2$
			& $0.154$ & $44.7 \pm 1.6$
			& -0.000 & +0.0 \\
			CoMo \cite{yang2025learning}
			& $0.181$ & $74.1 \pm 1.8$
			& $0.125$ & $87.6 \pm 1.3$
			& $0.221$ & $80.8 \pm 2.7$
			& $0.153$ & $49.9 \pm 1.8$
			& +0.002 & +0.5 \\
			CoMo w/ OF 
			& $0.172$ & $76.2 \pm 1.5$
			& \cellcolor{red!10}{$\mathbf{0.129}$} & \cellcolor{red!10}{$\mathbf{89.7 \pm 
					1.2}$}
			& $0.113$ & $82.6 \pm 2.4$
			& \cellcolor{red!10}{$\mathbf{0.139}$} & \cellcolor{red!10}{$\mathbf{57.9 \pm 1.8}$}
			& -0.030 & +4.0 \\
			LAOF 
			& \cellcolor{red!10}{$\mathbf{0.111}$} & \cellcolor{red!10}{$\mathbf{82.5 \pm 2.3}$}
			& $0.082$ & $85.3 \pm 1.4$
			& \cellcolor{red!10}{$\mathbf{0.118}$} & \cellcolor{red!10}{$\mathbf{87.2 \pm 2.2}$}
			& $0.088$ & $52.0 \pm 1.7$
			& \cellcolor{red!10}{$\mathbf{-0.069}$} & \cellcolor{red!10}{$\mathbf{+4.2}$} \\
			\midrule
			LAOM-Action \cite{nikulin2025latent}
			& $0.108$ & $86.0 \pm 2.3$
			& $0.090$ & $91.1 \pm 1.5$
			& $0.127$ & $86.3 \pm 1.7$
			& $0.086$ & $61.6 \pm 2.3$
			& -0.066 & +8.7 \\
			LAOF-Action 
			& \cellcolor{red!10}{$\mathbf{0.076}$} & \cellcolor{red!10}{$\mathbf{88.2 \pm 1.5}$}
			& \cellcolor{red!10}{$\mathbf{0.064}$} & \cellcolor{red!10}{$\mathbf{95.9 \pm 1.3}$}
			& \cellcolor{red!10}{$\mathbf{0.081}$} & \cellcolor{red!10}{$\mathbf{88.6 \pm 1.6}$}
			& \cellcolor{red!10}{$\mathbf{0.068}$} & \cellcolor{red!10}{$\mathbf{63.7  \pm 1.9}$}
			& \cellcolor{red!10}{$\mathbf{-0.096}$} & \cellcolor{red!10}{$\mathbf{+11.5}$} \\
			\bottomrule
		\end{tabular}
		\label{table:libero}
	\end{minipage}
	
	\vspace{1.2em}
	\begin{minipage}{\linewidth}
		\centering
		\subfloat[Action-Free]{%
			\def\svgwidth{0.238\linewidth}	
			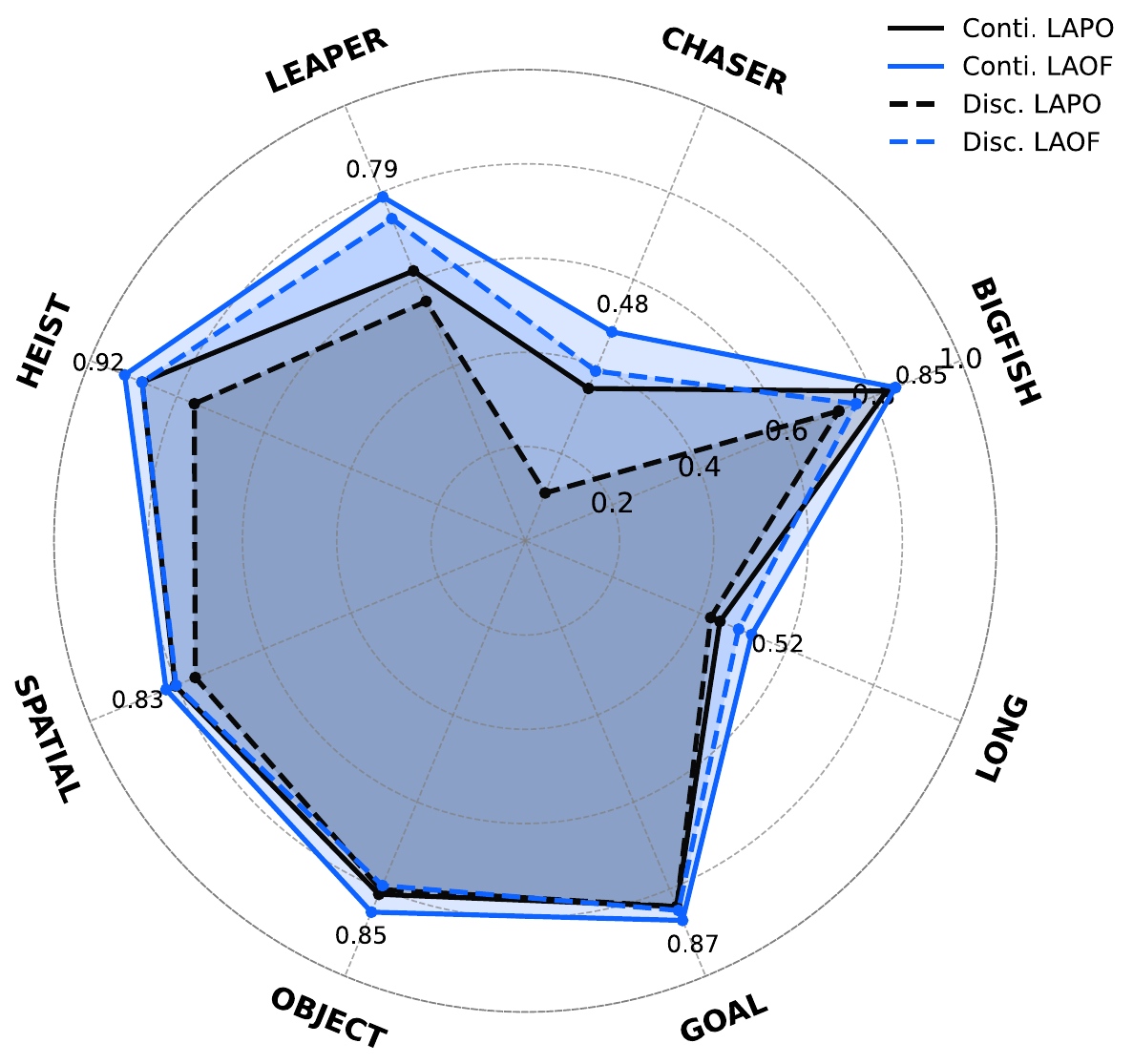
		}
		\hfill
		\subfloat[$1\%$ Action Ratio]{%
			\def\svgwidth{0.238\linewidth}
			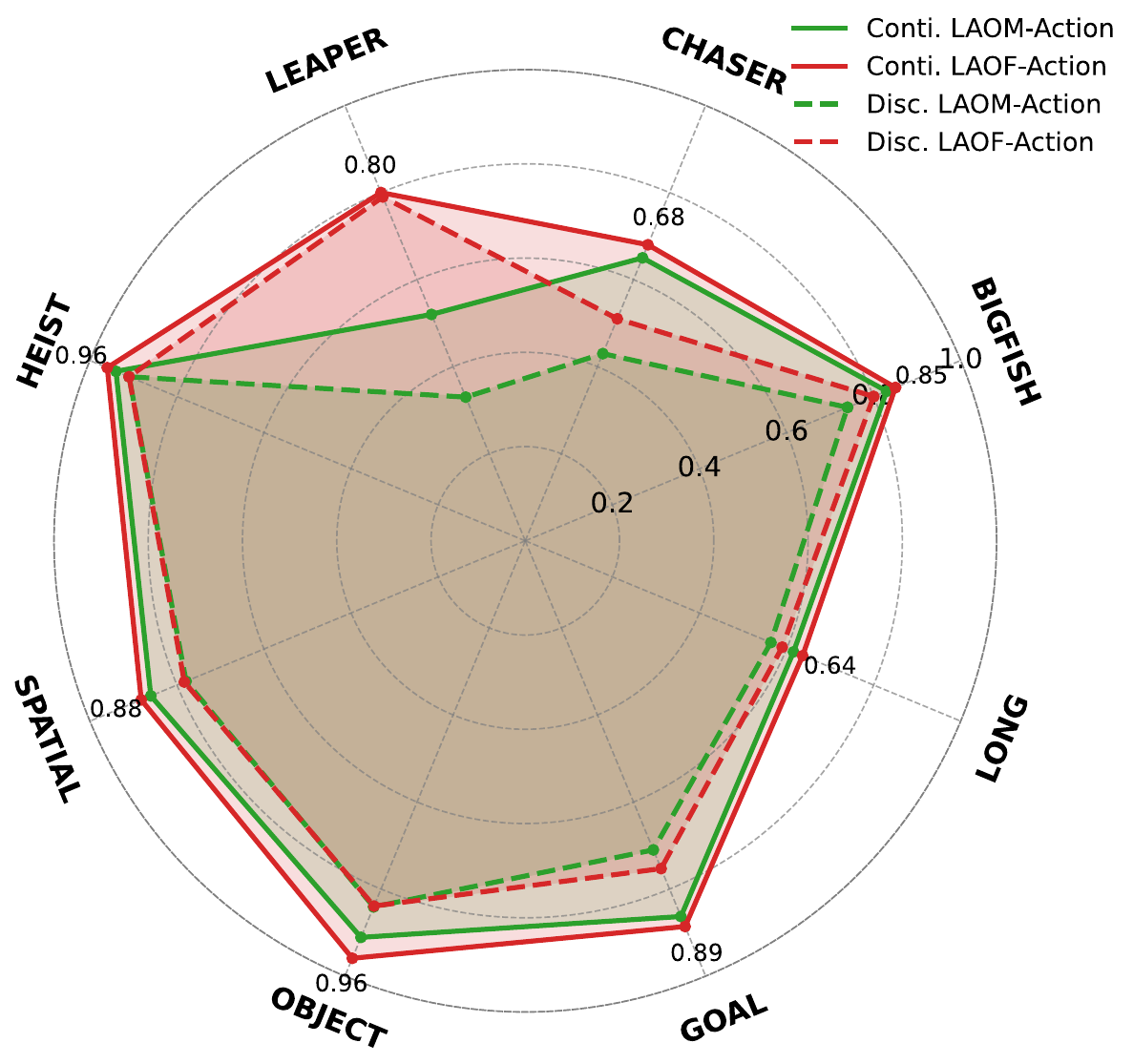
		}
		\hfill
		\subfloat[LIBERO]{%
			\def\svgwidth{0.238\linewidth}
			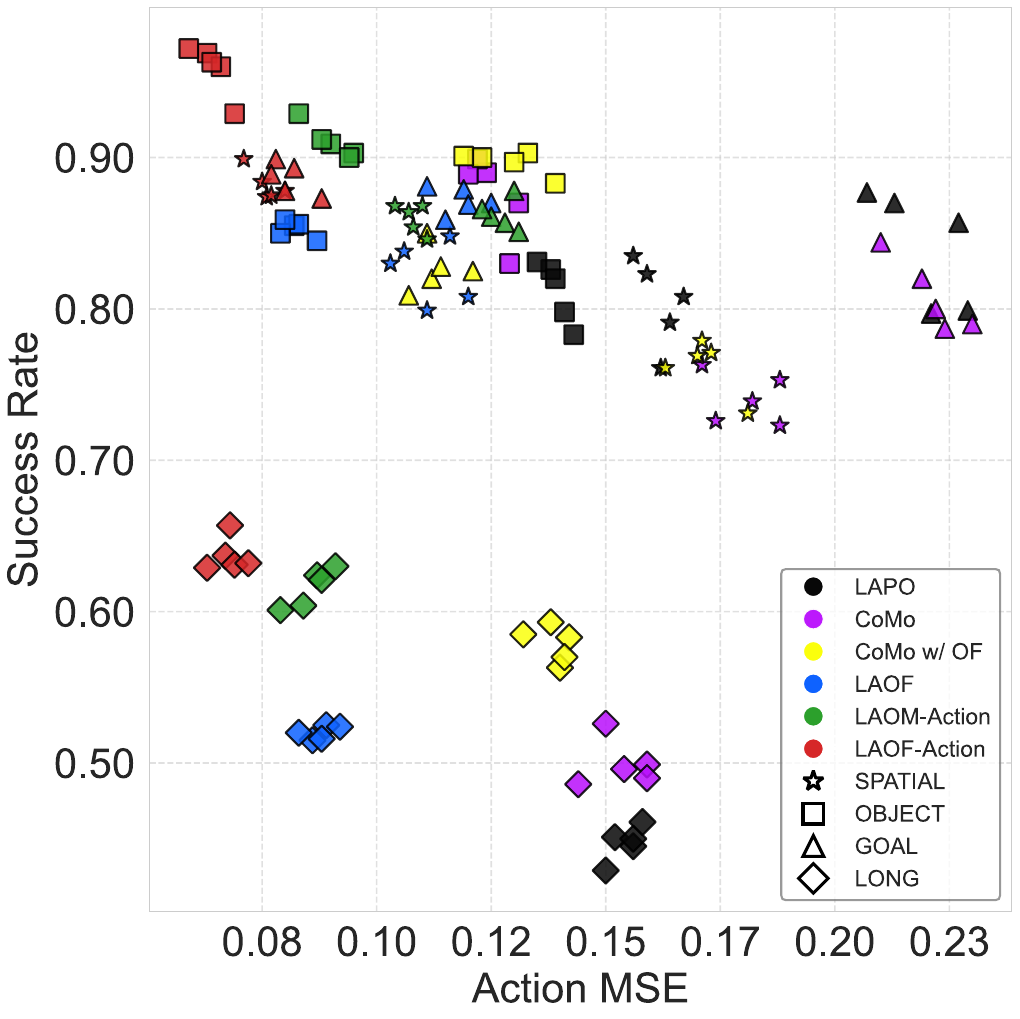
		\textsc{\textsc{\textsc{}}}		}
		\hfill
		\subfloat[PROCGEN]{%
			\def\svgwidth{0.238\linewidth}
			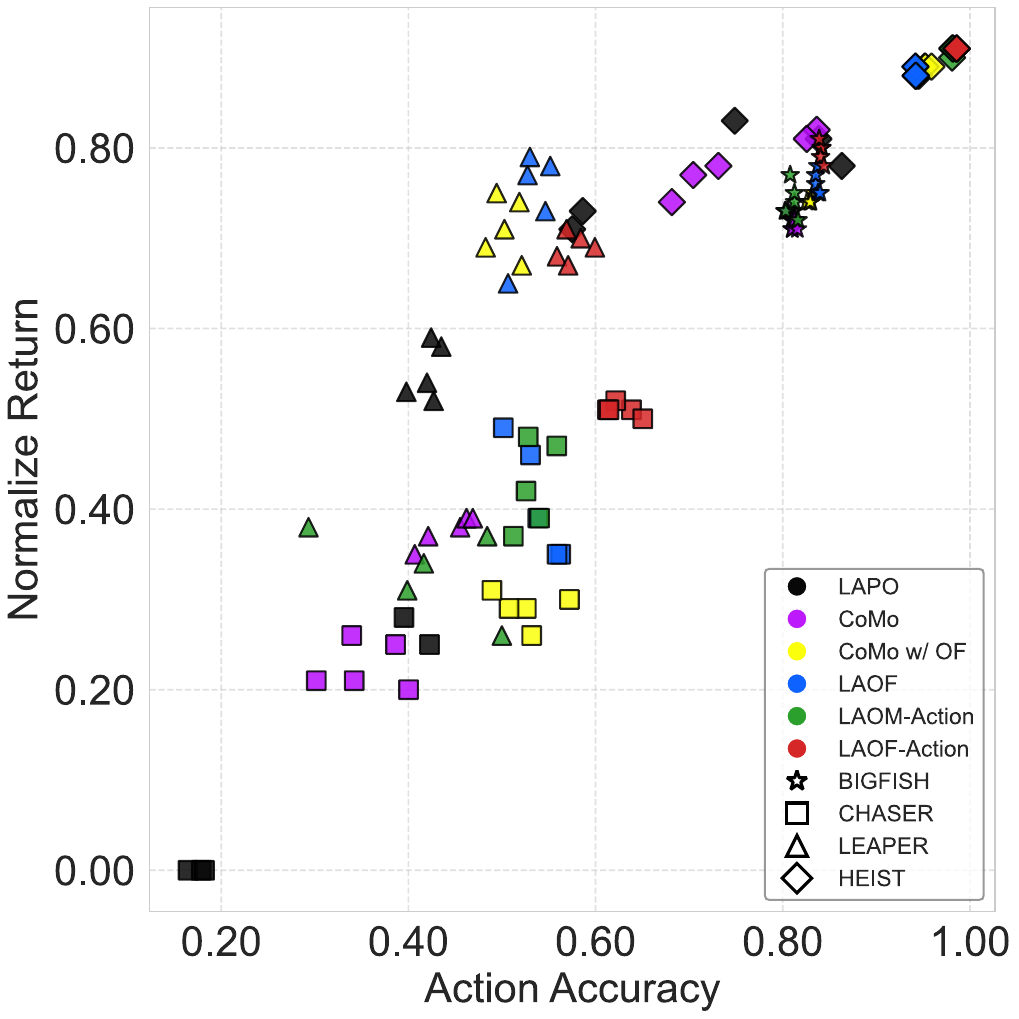
		}
		\caption{Subfigures (a) and (b) compare downstream task performance between continuous (solid lines) and discrete (dashed lines) latent action representations, using normalized episodic return for PROCGEN and success rate for LIBERO. The larger area under the solid lines compared to the dashed lines indicates that continuous representations outperform discrete ones across all downstream tasks (\textbf{Q1}). Subfigures (c) and (d) illustrate that our latent action evaluation metric is highly correlated with downstream task performance, with the mean Pearson correlation coefficient averaged across all tasks being 0.8288 for PROCGEN and –0.7311 for LIBERO.}
		\label{fig:radar_metrics}
	\end{minipage}
	
\end{figure*}

\subsection{Optical Flow in RGB Format}
\label{subsec:flow2rgb}
Given two consecutive observations $(o_t, o_{t+1})$, the optical flow $f_{\text{uv}, t} \in \mathcal{F}_{\text{uv}}$ represents the pixel-level motion from $o_t$ to $o_{t+1}$, where $\mathcal{F}_{\text{uv}} = \mathbb{R}^{H \times W \times 2}$. Each pixel in $f_{\text{uv}}$ is described by two channels, $(u,v)$, corresponding to the horizontal and vertical motion components, respectively. Since raw optical flow is not directly compatible with DINOv2’s input format, we follow the technique of VideoJAM \cite{chefer2025videojam} to convert it into RGB-formatted optical flow, i.e., $\mathcal{F}_{\text{uv}} \rightarrow \mathcal{F}_{\text{rgb}}$, enabling consistent processing alongside the original images. Specifically, each optical flow vector $(u, v)$ is first converted into polar coordinates, with the direction $\alpha = \operatorname{atan2}(v, u)$ mapped to the Hue channel and the magnitude $m = \sqrt{u^2 + v^2}$ mapped to the Saturation and Value channels. The resulting HSV image is then converted to an RGB-formatted optical flow $f_{\text{rgb}}$ using a standard color transformation. To capture motions of varying scales effectively, the magnitude $m$ is further normalized as follows,
\begin{equation}
	m_{\text{norm}} = \min\!\left(1.0,\; \frac{m}{\sigma \cdot \sqrt{H^2 + W^2}}\right),
\end{equation}
where $\sigma$ is a sensitivity factor that balances subtle and pronounced motions.

\subsection{Object-Centric Optical Flow}
To generate optical flow pseudo-labels, we use RAFT \cite{teed2020raft}, a pre-trained optical flow model that provides accurate and dense pixel-level motion estimates between consecutive observations, as illustrated in Fig.\ref{fig:of}. Optical flow naturally filters out static distractors and can be directly applied to several popular robot manipulation datasets\cite{o2024open, khazatsky2024droid, liu2023libero} without additional processing. In scenarios with dynamic distractors, such as pixel-based games \cite{cobbe2020leveraging, mnih2013playing, guss2019minerl}, action-irrelevant motion may introduce noise that hinders latent action learning, which requires a mechanism to obtain object-centric optical flow. Specifically, we employ LangSAM \cite{langSAM}, which generates an object-centric semantic mask for the current observation frame $o_t$ under the guidance of object-related textual prompts. The generated mask is then applied to the global optical flow, $f^{\mathrm{all}}_{\text{rgb},t}$, to filter out moving objects other than the agent, resulting in: $f^{\text{sam}}_{\text{rgb},t} = \text{mask}_t \odot f^{\mathrm{all}}_{\text{rgb},t}$, where $\odot$ denotes element-wise weighting, retaining only the optical flow features within the mask. The final supervision signal, $f_{\text{rgb},t} = \{f^{\mathrm{all}}_{\text{rgb},t}, f^{\text{sam}}_{\text{rgb},t}\}$, is selected adaptively according to the task scenario: in environments with static distractors, the global optical flow can be used directly, whereas in environments with dynamic distractors, the object-centric optical flow is adopted to capture motions relevant to the agent.

\begin{figure*}[t!]
	\centering
	
	\begin{minipage}{\linewidth}
		\centering
		\captionof{table}{Effect of discrete latent actions on downstream reinforcement learning performance on PROCGEN. Acc. denotes the accuracy of action classification ($\%$), Return denotes the average normalized episodic return over 1000 trials.}
		\label{tab:action_classification}
		\small 
		\setlength{\tabcolsep}{1pt}       
		\renewcommand{\arraystretch}{0.9} 
		\begin{tabular}{lcccccccccccc}
			\toprule
			\multirow{2}{*}{Method} 
			& \multicolumn{2}{c}{BIGFISH} 
			& \multicolumn{2}{c}{CHASER} 
			& \multicolumn{2}{c}{LEAPER} 
			& \multicolumn{2}{c}{HEIST} 
			& \multicolumn{2}{c}{Avg. Impr.} \\
			\cmidrule(lr){2-3} \cmidrule(lr){4-5} \cmidrule(lr){6-7} \cmidrule(lr){8-9} \cmidrule(lr){10-11}
			& Acc.~($\uparrow$) & Return ($\uparrow$) & Acc.~($\uparrow$) & Return ($\uparrow$) & Acc.~($\uparrow$) & Return  ($\uparrow$)& Acc.~($\uparrow$) & Return ($\uparrow$) & Acc.~($\uparrow$) & Return ($\uparrow$) \\
			\midrule
			LAPO \cite{schmidt2023learning} 
			& $80.98 \pm 0.28$ & $0.72$
			& $26.87 \pm 12.86$ & $0.11$
			& $40.09 \pm 1.40$ & $0.55$
			& $72.23 \pm 13.61$ & $0.76$
			& +0.00 & +0.00 \\
			CoMo \cite{yang2025learning}
			& $81.02 \pm 0.49$ & $0.72$
			& $35.39 \pm 3.96$ & $0.23$
			& $44.29 \pm 2.73$ & $0.56$
			& $75.58 \pm 7.09$ & $0.78$
			& +4.03 & +0.04 \\
			CoMo w/ OF 
			& $82.91 \pm 0.24$ & $0.74$
			& $52.53 \pm 3.10$ & $0.29$
			& $50.38 \pm 1.63$ & $0.71$
			& \cellcolor{red!10}{$\mathbf{95.05 \pm 0.59}$} & \cellcolor{red!10}{$\mathbf{0.89}$}
			& +15.18 & +0.12 \\
			LAOF 
			& \cellcolor{red!10}{$\mathbf{83.71 \pm 0.20}$} & \cellcolor{red!10}{$\mathbf{0.76}$}
			& \cellcolor{red!10}{$\mathbf{53.83 \pm 2.47}$} & \cellcolor{red!10}{$\mathbf{0.39}$}
			& \cellcolor{red!10}{$\mathbf{53.23 \pm 1.78}$} & \cellcolor{red!10}{$\mathbf{0.74}$}
			& $94.19 \pm 0.02$ & $0.88$
			& \cellcolor{red!10}{$\mathbf{+16.20}$} & \cellcolor{red!10}{$\mathbf{+0.16}$} \\
			\midrule
			LAOM-Action \cite{nikulin2025latent}
			& $81.05 \pm 0.51$ & $0.74$
			& $52.95 \pm 1.96$ & $0.43$
			& $41.85 \pm 8.22$ & $0.33$
			& $98.14 \pm 0.04$ & $0.91$
			& +13.46 & +0.07 \\
			LAOF-Action 
			& \cellcolor{red!10}{$\mathbf{84.13 \pm 0.18}$} & \cellcolor{red!10}{$\mathbf{0.80}$}
			& \cellcolor{red!10}{$\mathbf{62.75 \pm 1.63}$} & \cellcolor{red!10}{$\mathbf{0.51}$}
			& \cellcolor{red!10}{$\mathbf{57.64 \pm 1.57}$} & \cellcolor{red!10}{$\mathbf{0.79}$}
			& \cellcolor{red!10}{$\mathbf{98.57 \pm 0.01}$} & \cellcolor{red!10}{$\mathbf{0.91}$}
			& \cellcolor{red!10}{$\mathbf{+20.73}$} & \cellcolor{red!10}{$\mathbf{+0.22}$} \\
			\bottomrule
		\end{tabular}
		\label{table:procgen}
	\end{minipage}
	
	\vspace{0.7em}
	
	\begin{minipage}{\linewidth}
		\centering
		\subfloat[BIGFISH]{%
			\def\svgwidth{0.232\linewidth}
			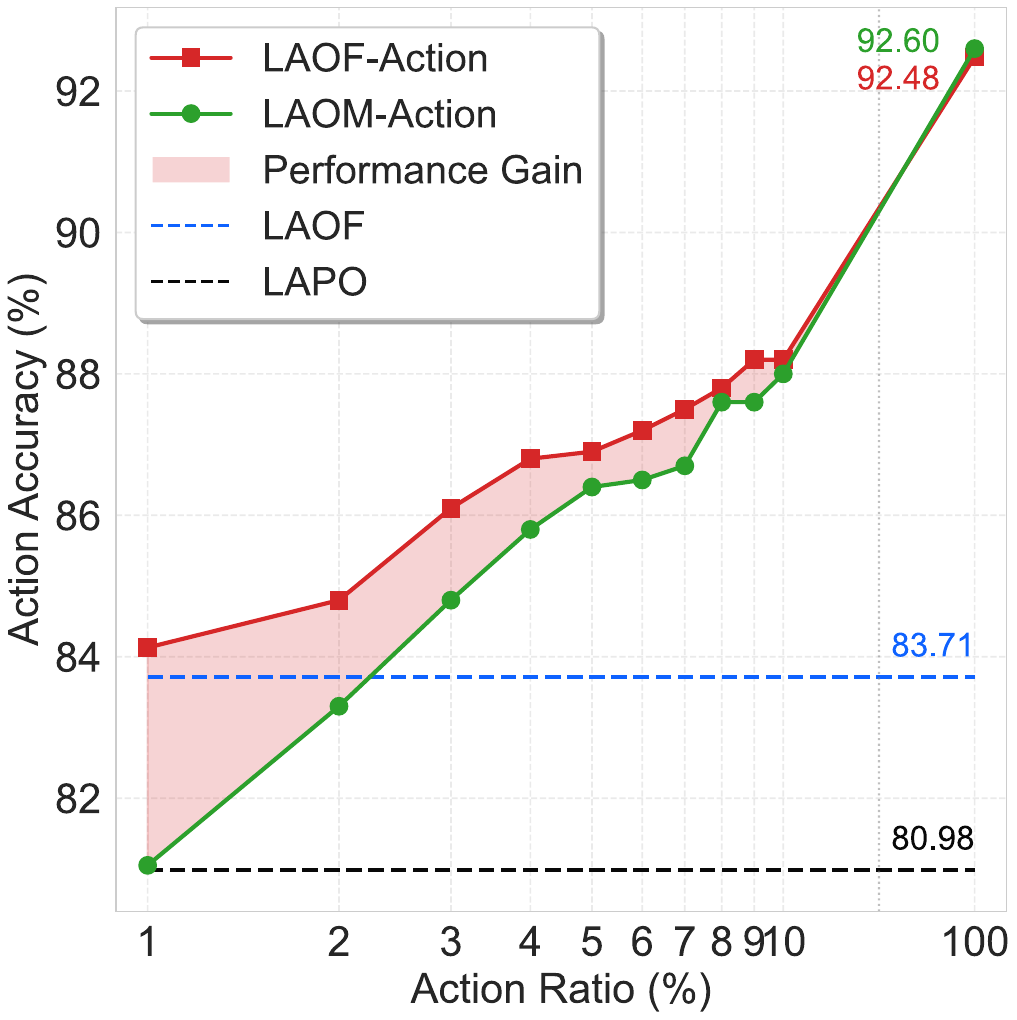
		}
		\hfill
		\subfloat[CHASER]{%
			\def\svgwidth{0.245\linewidth}
			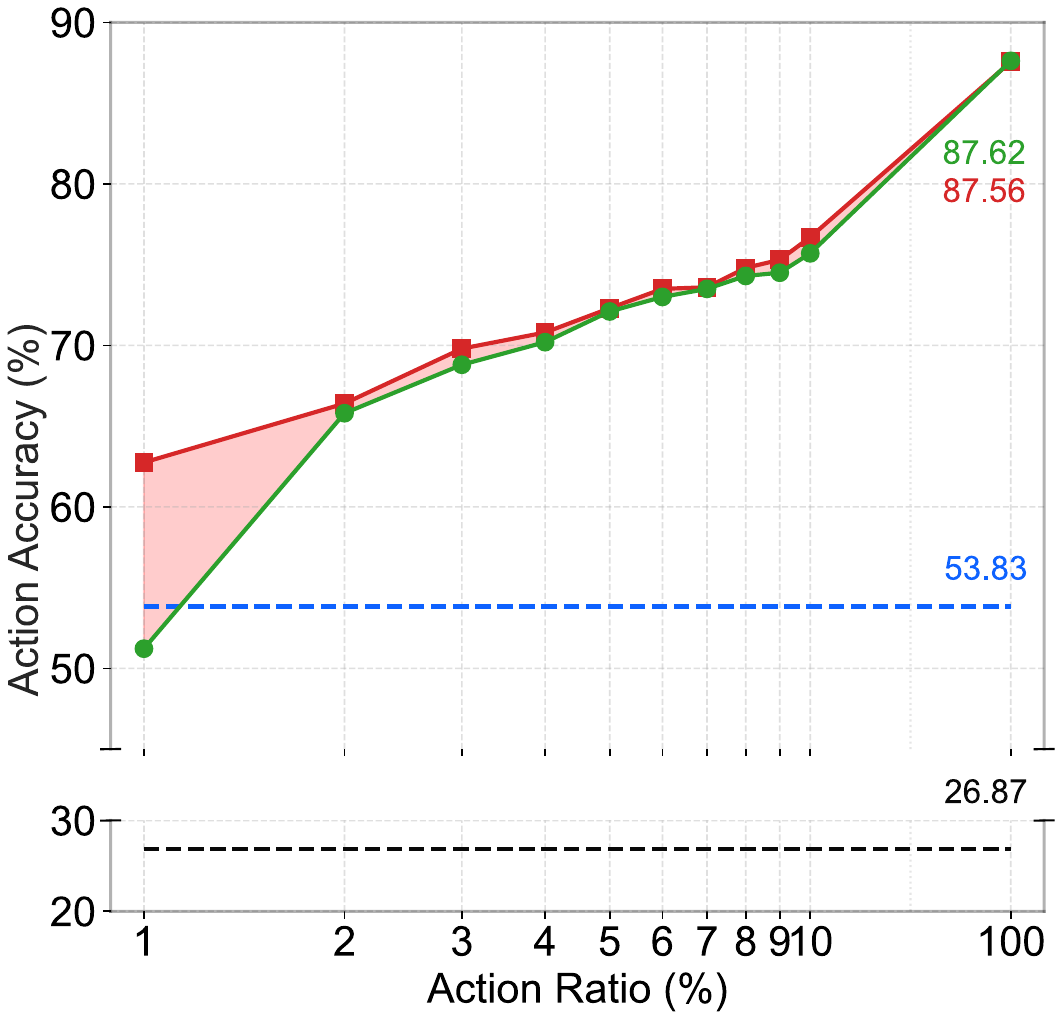
		}
		\hfill
		\subfloat[LEAPER]{%
			\def\svgwidth{0.232\linewidth}
			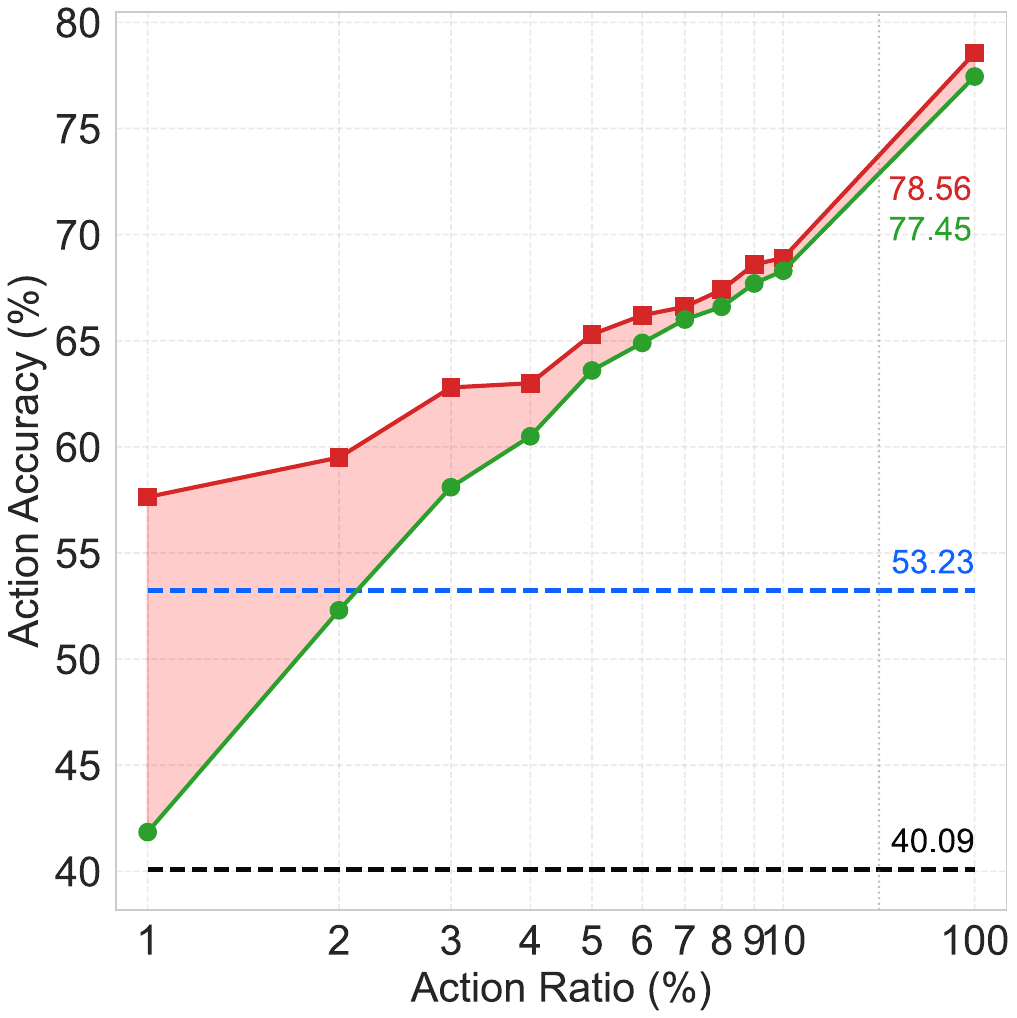
		}
		\hfill
		\subfloat[HEIST]{%
			\def\svgwidth{0.240\linewidth}
			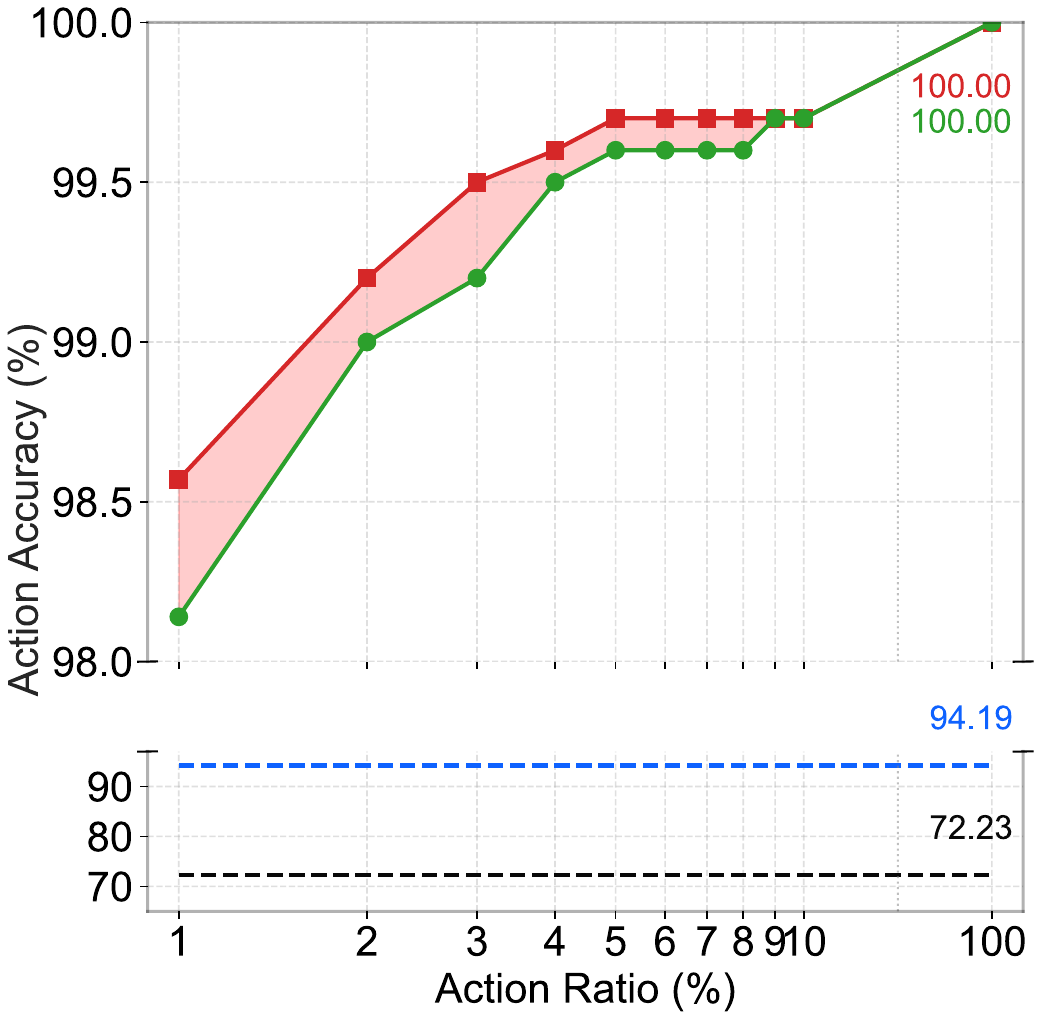
		}
		\caption{Evaluation of discrete latent action quality across different action ratios on PROCGEN. Action ratio indicates the proportion of action labels in the training dataset. A shared legend for all four subfigures is shown in subfigure (a). Optical flow constraints significantly improve the training stability of baselines and the quality of learned representations, combined with Fig.\ref{fig:traing_process}. Notably, under the extreme 1$\%$ action ratio, LAOF without any action supervision matches or even outperforms the action-supervised LAOM-Action.}
		\label{fig:action_ratio}
	\end{minipage}
	
\end{figure*}

\section{Experiments}
Our experiments aim to address the following questions:

(\textbf{Q1}) Which form of representation is more effective: continuous actions or discrete actions?

(\textbf{Q2}) Does introducing optical flow constraints enhance the quality of learned representations and improve downstream task performance?

(\textbf{Q3}) Up to what proportion of labeled actions in the training dataset does optical flow continue to provide performance gains over the baseline?

(\textbf{Q4}) To what extent can optical flow serve as a substitute for action labels under limited supervision?

(\textbf{Q5}) What is the optimal architecture for introducing optical flow constraints?

\subsection{Experimental Setup}

\noindent\textbf{Benchmark.} Our experimental setup includes both continuous and discrete control tasks, involving downstream imitation learning and reinforcement learning. For all tasks, the dimensionality of the latent action is set to 128, and each task is evaluated using 5 random seeds.

\textbullet{} LIBERO \cite{liu2023libero}: This benchmark is designed to investigate knowledge transfer in multitask robot learning. We adopt four task suites: SPATIAL, OBJECT, GOAL and LONG. Each suite contains 10 tasks, with 50 human-teleoperated demonstrations per task, 45 of which are used for the training set and the remaining 5 for the test set. Since all LIBERO tasks share the same environment, we aggregate data from all tasks into a unified training set to fully exploit the environment dynamics. Instead of building a powerful VLA model, we simply extend all baselines by incorporating language instructions to construct a multi-task model. This is because the optimal architectures for continuous and discrete representations remain uncertain when fine-tuning the base vision-language model, and we seek to prevent additional factors from affecting the ablation study. Moreover, omitting the fine-tuning step allows us to directly evaluate the effectiveness of optical flow constraints in enhancing representation learning and improving downstream imitation performance. The training process consists of three stages: a pre-training stage of 20 epochs, a distillation stage of 5 epochs, and a fine-tuning stage of 5 epochs.

\textbullet{} PROCGEN \cite{cobbe2020leveraging}: It is a procedurally generated discrete control reinforcement learning benchmark that incorporates both static distractors and dynamic distractors induced by stochastic effects, which are not driven by the agent’s actions. We adopt four tasks: BIGFISH, CHASER, LEAPER, HEIST. Among these environments, the first three contain dynamic distractors, while the last involves only static ones. Since these tasks are set in different environments, we train separate models for each task. The training and test sets contain approximately 2M and 0.2M frames, respectively. All data are collected using an expert policy trained with PPO over 50M frames. The training process consists of three stages: a pre-training stage of 50K steps, a distillation stage of 60K steps, and a reinforcement learning stage of 4M steps. During the reinforcement learning stage, a hybrid policy head is employed, comprising a policy head fine-tuned via imitation learning for 3 epochs and an initialized policy head that is updated through reinforcement learning.

\noindent\textbf{Baselines.}
We select several typical latent action learning methods as baselines for comparison.

\textbullet{} LAPO \cite{schmidt2023learning}: An original baseline that jointly trains inverse and forward dynamics models through a reconstruction objective, without explicit motion constraints.

\textbullet{} CoMo \cite{yang2025learning}: It builds upon LAPO by using a temporal-difference strategy in the IDM input, where the next frame is replaced with inter-frame differences to encourage latent actions to capture temporal variations between frames.

\textbullet{} LAOM-Action \cite{nikulin2025latent}: It is an action-supervised method that incorporates an additional action decoder mapping latent actions to physical actions, enforcing consistency between the learned representations and physical actions.

\noindent\textbf{Metrics.}\label{sec:metrics} We train the latent action decoder for 3 epochs and use its performance to evaluate the quality of the latent actions learned during training. For discrete latent actions, the evaluation metric is classification accuracy (top-1): $\mathrm{Acc} = \frac{1}{M} \sum_{i=1}^{M} \mathbbm{1}[\hat{a}_i = a_i]$, where $a_i$ is the ground-truth action, $\hat{a}_i$ is the predicted action, and $\mathbbm{1}[\cdot]$ is the indicator function, which equals 1 if the condition inside is true and 0 otherwise. For continuous latent actions, the metric is the mean squared error (MSE) between the predicted and ground-truth actions: $\mathrm{MSE} = \frac{1}{M} \sum_{i=1}^{M} \|\hat{a}_i - a_i\|_2$.

\subsection{Conclusion}

\begin{figure}[t!]
	\centering
	\def\svgwidth{0.99\linewidth}
	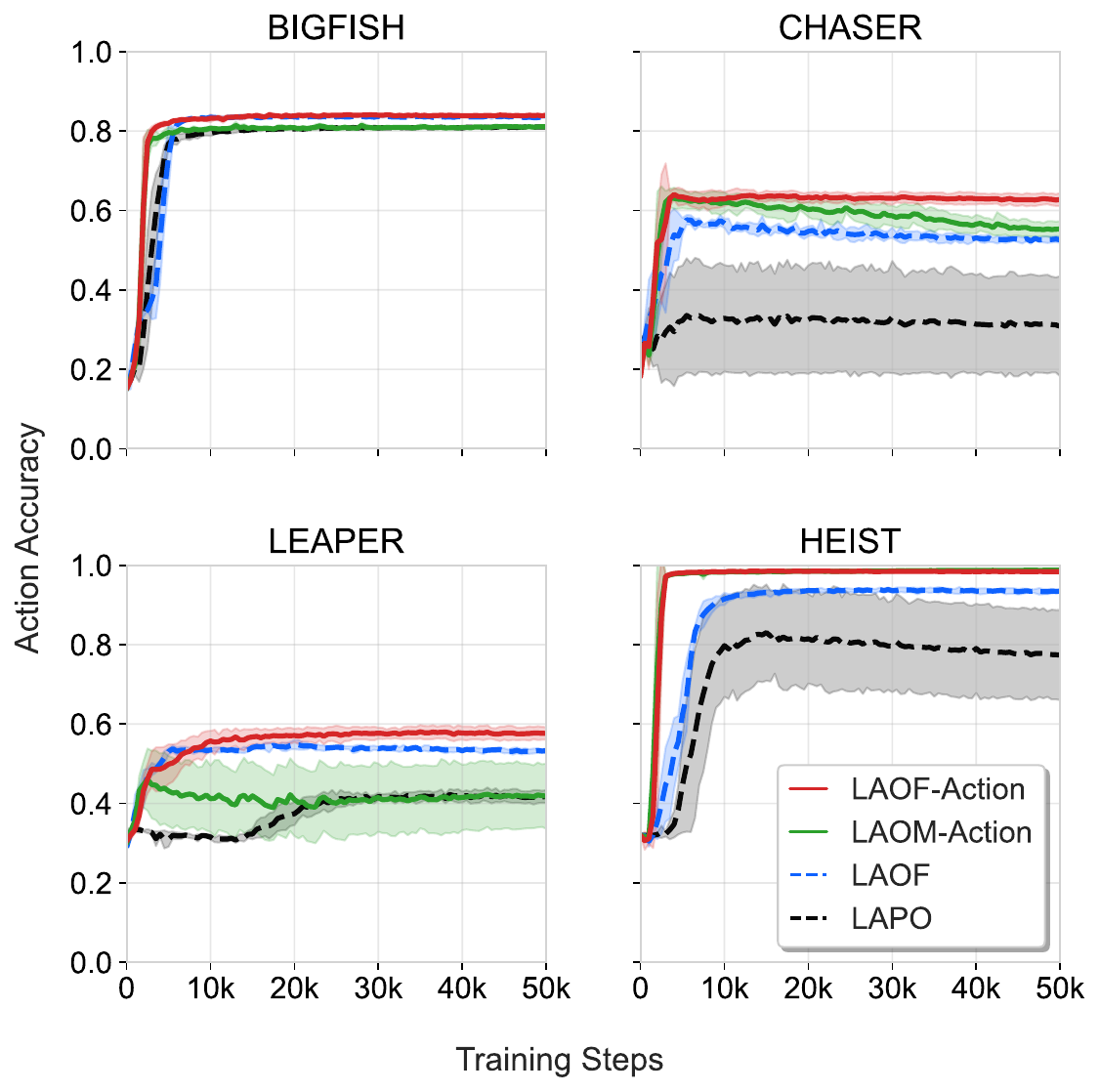
	\caption{Comparison of stability and overfitting among different methods, where solid lines represent unsupervised methods and dashed lines represent action-supervised methods. LAOM-Action and LAOF-Action are evaluated at a 1$\%$ action ratio.}
	\label{fig:traing_process}
\end{figure}

The results shown in Fig.~\ref{fig:radar_metrics} validate that our proposed evaluation metric effectively assesses the quality of latent action representations. Combined with Tables~\ref{table:libero} and~\ref{table:procgen}, optical flow constraints improve success rates by 4.2$\%$ for unsupervised methods and 11.5$\%$ for action-supervised methods on the LIBERO benchmark. On the PROCGEN benchmark, they increase normalized episodic rewards by 16$\%$ and 22$\%$, respectively. Together, these results indicate that incorporating optical flow constraints during pre-training substantially enhances the quality of latent action representations across both unsupervised and action-supervised methods, leading to improved performance on downstream imitation learning and reinforcement learning tasks (\textbf{Q2}). Moreover, we find that replacing the IDM’s future frame input with inter-frame differences, as done in CoMo \cite{yang2025learning}, does not lead to consistent performance improvements, as evidenced by performance degradation in the SPATIAL and GOAL tasks. Although inter-frame or encoded feature differences can capture ‘changes’, they do not provide a faithful approximation of physical motion or actions, as they lack pixel-level direction, magnitude, and locality, and cannot represent true physical movements like optical flow.

To investigate the potential performance gain that optical flow constraints can provide to action-supervised methods at different action ratios, we evaluated the action accuracy of different methods as the action ratio increases, as illustrate in Fig.\ref{fig:action_ratio}. We find that optical flow constraints consistently enhance the quality of latent action representations for baseline as the action ratio increases up to 10$\%$ (\textbf{Q3}). Beyond this point, the positive effects diminish, and at an action ratio of 100$\%$, performance degradation is even observed in the BIGFISH and CHASER environments, likely due to noise in the optical flow pseudo-labels. Accordingly, we suggest applying optical flow constraints when the dataset contains less than 10$\%$ action-labeled data. When the proportion exceeds this threshold, their use should be considered on a case-by-case basis. If optical flow constraints achieve performance comparable to action supervision alone, we recommend using the latter, as it provides a simpler network architecture, faster training, and reduces the risk of noise from low-quality optical flow pseudo-labels.

As shown in Fig.~\ref{fig:training}, the original LAPO demonstrates highly unstable performance with large standard deviations in the CHASER and HEIST tasks, indicating that, without explicit motion constraints, the learned latent actions are easily influenced by action-irrelevant distractors such as moving objects or background changes. Under the extreme condition of a 1$\%$ action ratio, LAOM-Action also exhibits training instability and overfitting in the LEAPER and CHASER tasks, even with action supervision, as sparse action supervision makes the model prone to spurious correlations in visual features. By introducing optical flow constraints, LAOF-Action shows significantly improved stability across all tasks. The learning curves show faster convergence, smaller variance, and higher final action accuracy, indicating that optical flow serves as an effective supervision signal to regularize latent action learning. Notably, LAOF without any action supervision achieves performance comparable to or even outperforms that of LAOM-Action trained with sparse action labels (1$\%$), highlighting the strong capability of flow-based constraints in capturing meaningful physical dynamics and mitigating overfitting (\textbf{Q4}).

\section{Discussion}
To investigate how the form and placement of optical flow constraints affect latent action learning, we conduct a focused set of ablations that vary both the constraint structure and its coupling with the FDM. Specifically, we examine the following structures:

\begin{itemize}
\item \textbf{LAOF:} It builds upon LAPO and introduces a dedicated decoder that directly maps latent actions to optical flow.

\item \textbf{LAOF-FlowFDM:} It integrates the flow decoder into the FDM to simplify the architecture, enabling the model to jointly predict both optical flow and the next state.

\item \textbf{LAOF-Only ($z_t$):} By removing the FDM and retaining a dedicated flow decoder, this variant examines whether latent actions can be learned solely through optical flow constraints to evaluate the necessity of the FDM.

\item \textbf{LAOF-Only ($z_t, s_t$):} The flow decoder takes both the latent action and the current state as input. This design assesses the effect of applying optical flow constraints to both state and latent action, testing whether such constraints should be integrated into the FDM.

\item \textbf{LAOF-AE:} It conducts optical flow autoencoding to explore whether this mechanism can independently yield a visual encoder for embodied foundation models.

\end{itemize}

The experimental results reveal a clear performance hierarchy: LAOF $>$ LAOF-AE $>$ LAOF-Only($z_t$) $>$ LAOF-FlowFDM $>$ LAOF-Only($z_t, s_t$), from which several insights emerge. First, LAOF-Only($z_t$) removes the FDM relative to LAOF and exhibits reduced performance, suggesting that the FDM provides structural context beneficial for latent action learning. Second, naively integrating the flow decoder into the FDM, as in LAOF-FlowFDM, to simplify the architecture proves suboptimal. The performance of LAOF-Only($z_t, s_t$) further confirms this, falling below that of LAOF-Only($z_t$) and even underperforming LAOF-FlowFDM. This effect may arise from applying optical flow constraints simultaneously to both latent actions and the current state, which allows the model to form direct dependencies from the state to optical flow, preventing the motion constraints from fully regularizing latent action learning. Finally, LAOF achieves the best performance, demonstrating that introducing optical flow constraints with a dedicated decoder provides strong and direct physical supervision for latent actions (\textbf{Q5}).

Moreover, we explored a simplified optical flow autoencoding variant, LAOF-AE, whose competitive performance demonstrates that optical flow supervision alone provides strong prior information about physical motions for learning latent actions. These findings suggest that training optical flow encoders through unsupervised learning could be an effective strategy for developing general-purpose embodied foundation models.

\begin{table}[t!]
	\centering
	\caption{Average performance improvement of LAOF variants over LAPO across all LIBERO and PROCGEN tasks.}
	\label{tab:action_classification_libero}
	\renewcommand{\arraystretch}{1}
	\setlength{\tabcolsep}{3pt} 
	\resizebox{0.98\linewidth}{!}{%
		\begin{tabular}{lcccc}
			\toprule
			\multirow{2.5}{*}{\centering Method} & \multicolumn{2}{c}{LIBERO} & \multicolumn{2}{c}{PROCGEN} \\
			\cmidrule(lr){2-3} \cmidrule(lr){4-5}
			& MSE ($\downarrow$) & Succ. ($\uparrow$) & Acc. ($\uparrow$) & Return ($\uparrow$) \\
			\midrule
			LAPO & -0.000 & +0.0 & +0.00 & +0.00 \\
			LAOF-FlowFDM & -0.046 & +3.3 & +11.95 & +0.13 \\
			LAOF-Only($z_t$) & -0.053 & +3.6 & +13.2 & +0.14 \\
			LAOF-Only($z_t,s_t$) & -0.012 & +0.8 & +6.26 & +0.03 \\
			LAOF-AE & -0.060 & +3.9 & +13.31 & +0.14 \\
			LAOF & \cellcolor{red!10}{$\mathbf{-0.069}$} & \cellcolor{red!10}{$\mathbf{+4.2}$} & \cellcolor{red!10}{$\mathbf{+16.20}$} & \cellcolor{red!10}{$\mathbf{+0.16}$} \\
			\bottomrule
	\end{tabular}}
	\label{fig:training}
\end{table}
\section{Limitations and Future Work}
This work primarily investigates whether agent optical flow can improve the robustness of latent action learning to action-irrelevant distractors, and what is the optimal architecture for introducing optical flow constraints. To extract agent optical flow, we leverage LangSAM \cite{langSAM} to obtain object segmentations guided by text prompts. While this approach is effective, text-driven segmentation is inherently challenging to precisely regulate, and the resulting masks often fail to accurately capture the target agents. Future work could incorporate click-based or interactive segmentation techniques to obtain more accurate agent masks, facilitating high-quality annotations for large-scale datasets.

Moreover, our approach currently applies only to “eye-to-hand” scenarios, where the environment is static and the agent moves within the visual field. In contrast, “eye-in-hand” settings feature cameras mounted on the agent, with the agent centered while the environment moves. Such data are typically captured from a wrist-mounted perspective, as in the current Universal Manipulation Interface (UMI) dataset \cite{chi2024universal, liu2025fastumi}. Extending LAOF to these scenarios requires disentangling agent and environmental motion. Future work will incorporate egocentric human videos, explore cross-embodiment settings, and address challenges such as camera shake and viewpoint variations, aiming to develop a general LAOF framework for diverse, multi-view, large-scale datasets and capable of robustly learning latent actions across heterogeneous visual configurations.

\section*{Acknowledgment}
We thank Shanghai Institute for Mathematics and Interdisciplinary Sciences (SIMIS) for their financial support. This work was funded by SIMIS under Grant No. SIMIS-ID-2025-RB and by the National Natural Science Foundation of China under Grant No. 62422316. The authors are grateful for the resources and facilities provided by SIMIS, which were essential for the completion of this work.

\bibliographystyle{ieeenat_fullname}
\bibliography{reference}

\clearpage
\appendix
\onecolumn 

\begin{center}
	{\LARGE \textbf{Appendix}}
\end{center}
\vspace{1em} 

\section{Additional Experimental Results}
\label{Appendix:eperimental_results}

\subsection{Visualization of Optical Flow}
\begin{figure}[h!]   
	\centering
	\includegraphics[width=\textwidth]{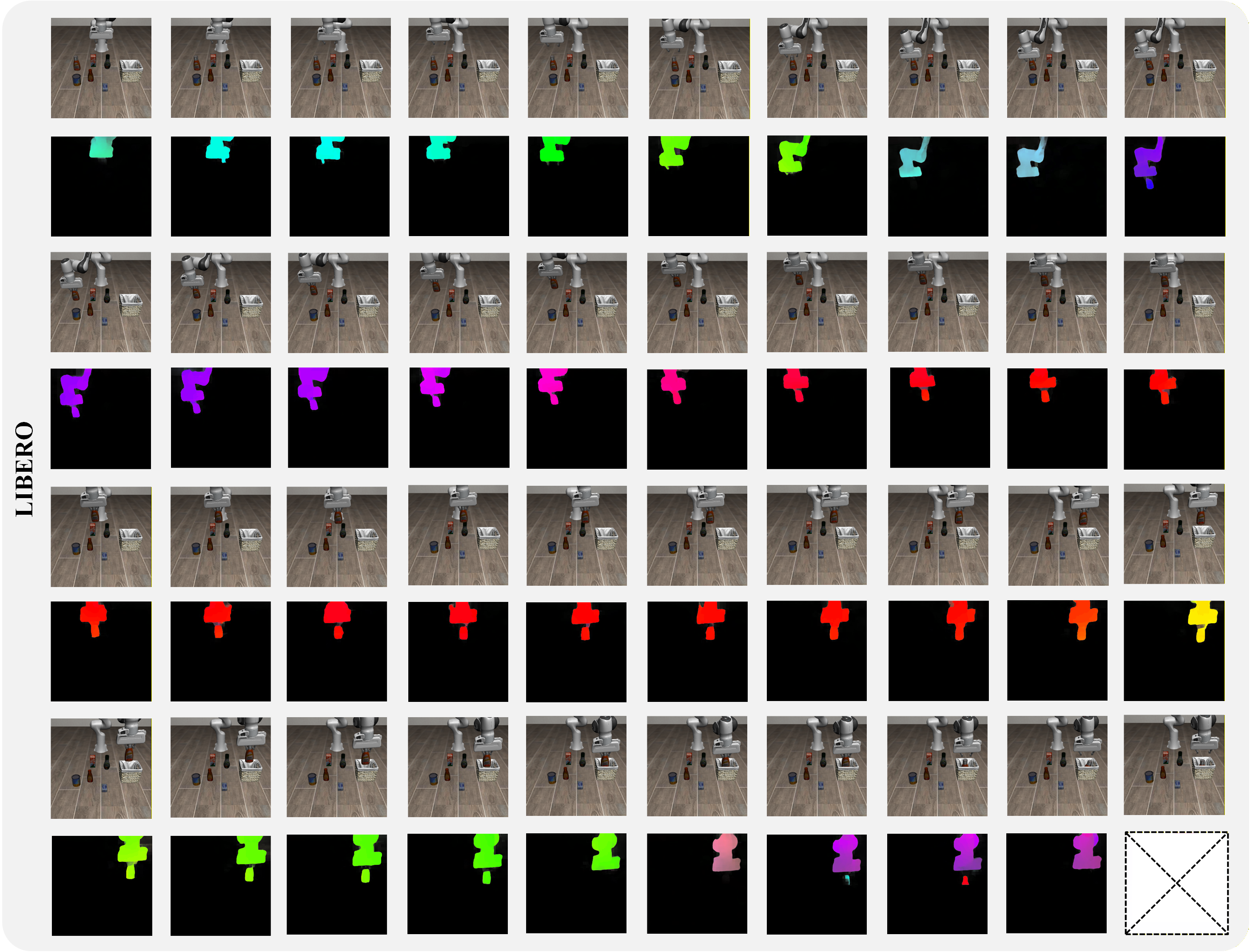}
	\caption{Visualization of optical flow for the LIBERO task: \textit{``pickup the ketchup and place it in the basket"}. To illustrate the workflow, we sampled 40 frames from the full 147 frames sequence at intervals of 3 frames, preserving the temporal order.}
	\label{fig:object_of}
\end{figure}

As illustrated in Fig.~\ref{fig:object_of}, we visualize the optical flow for a representative complete workflow of a task in LIBERO. SPATIAL tests reasoning about spatial relationships for accurate bowl placement. OBJECT evaluates generalization across varying object types within the same layouts. GOAL examines adaptive, goal-oriented behavior under varying objectives. LONG focuses on long-horizon tasks with multiple subgoals, assessing multi-step planning with heterogeneous objects and layouts.

Fig.~\ref{fig:bigfish_of} shows the workflows of four tasks on the PROCGEN benchmark, along with their corresponding raw and object-centric optical flow. The tasks are: BIGFISH, where the player grows by eating smaller fish while avoiding larger ones, receiving small rewards for eating and a large reward for becoming the biggest; CHASER, a maze navigation task in which the player collects green orbs while avoiding enemies, with power-ups that temporarily make enemies vulnerable; HEIST, where the player must steal a gem by collecting color-coded keys to unlock corresponding locks in a maze, with keys in possession displayed on-screen; and LEAPER, inspired by Frogger, in which the player crosses lanes of cars and hops on logs to traverse rivers, earning rewards for reaching the finish line.


\clearpage 

{\centering
\includegraphics[width=\textwidth]{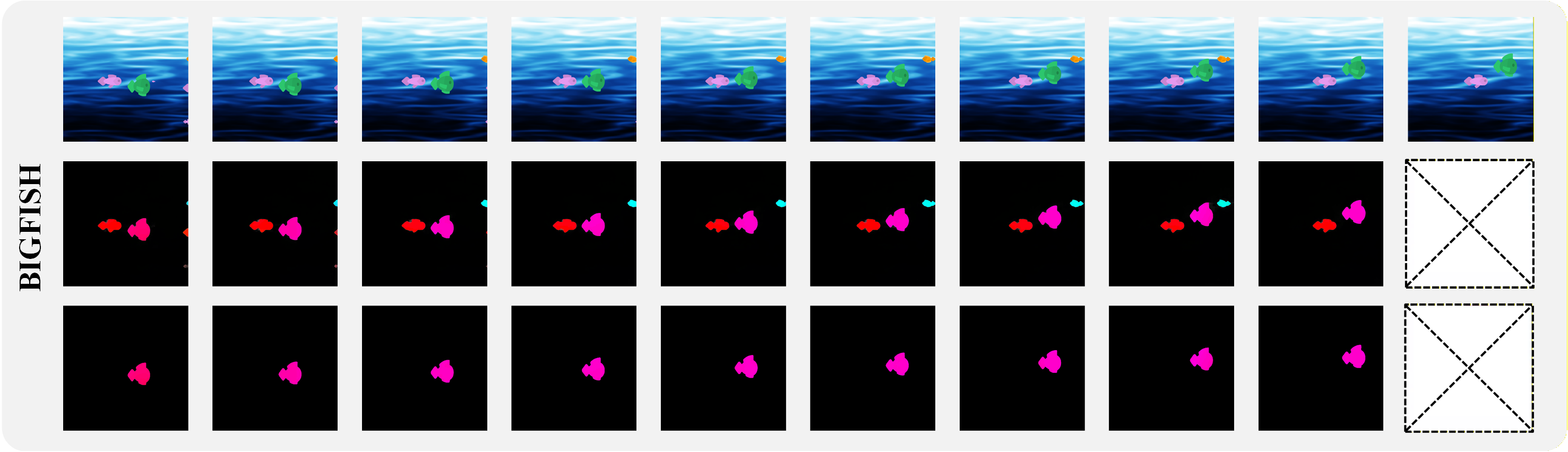}
\vspace{-0.8em} \\
\includegraphics[width=\textwidth]{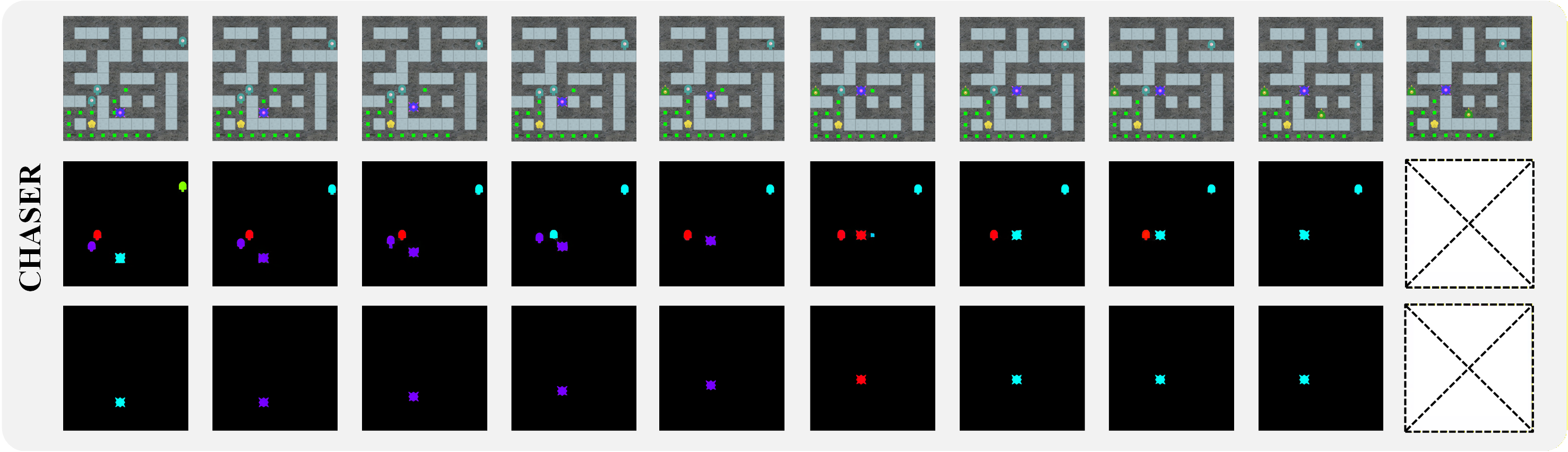}
\vspace{-0.8em} \\
\includegraphics[width=\textwidth]{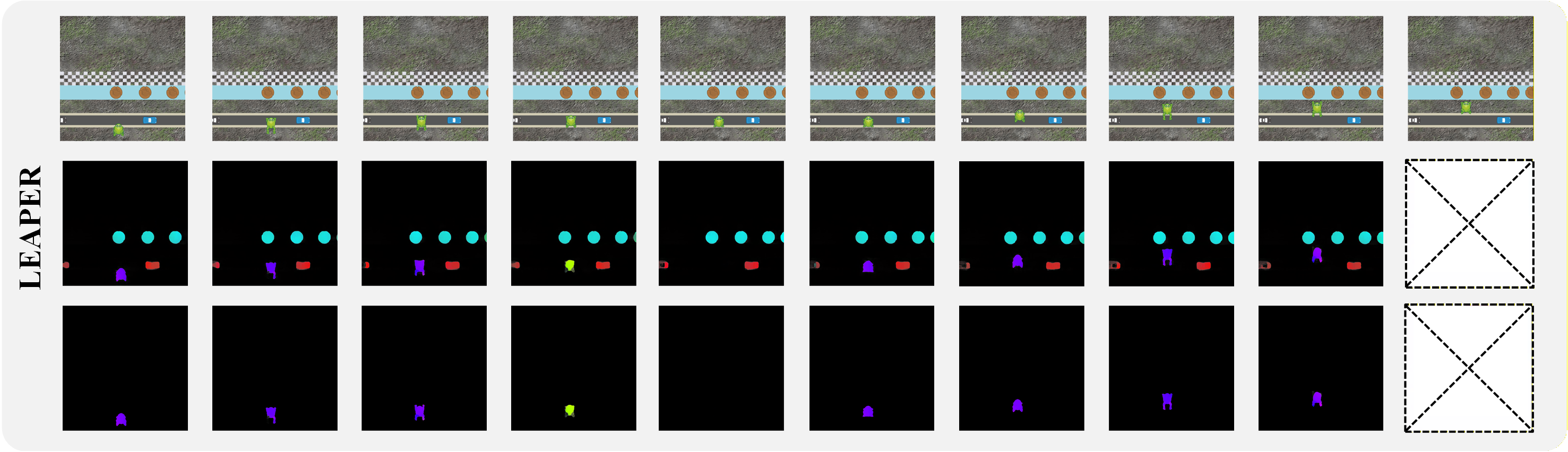}
\vspace{-0.8em} \\
\includegraphics[width=\textwidth]{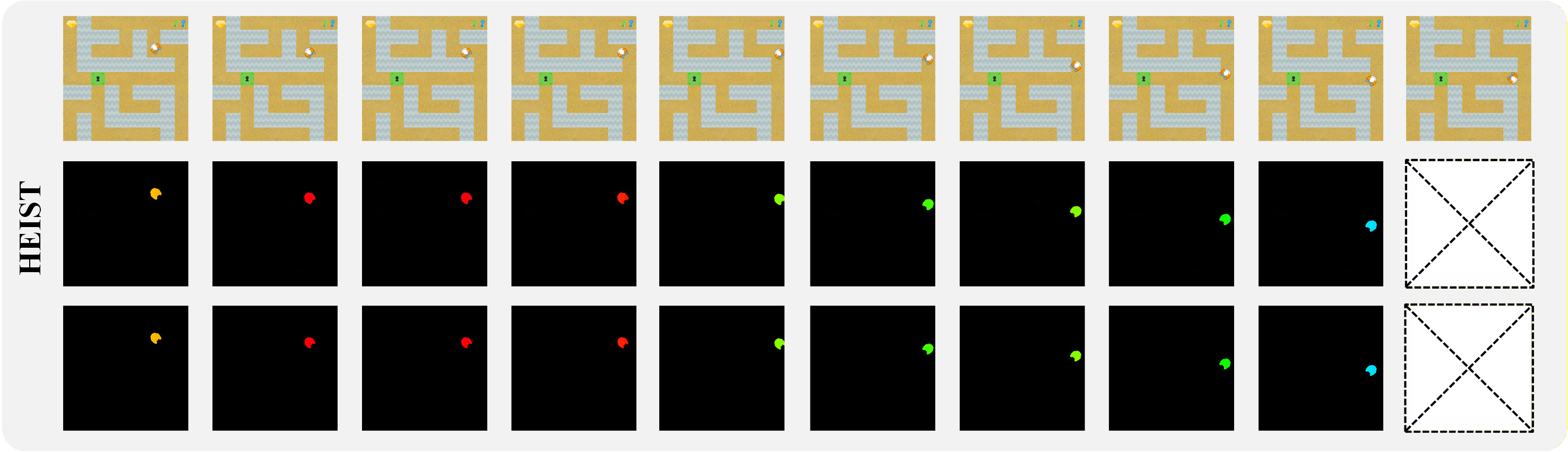}
\captionof{figure}{Visualization of optical flow for four tasks on the PROCGEN benchmark. For each task, 10 frames were sampled to illustrate the workflow, with each column representing a frame in temporal order. The first row shows the original images, while the second and third rows display the raw optical flow and the object-centric optical flow, respectively.}
\label{fig:bigfish_of}
}

\begin{figure}[h!]
	\centering
	\begin{minipage}[t]{\columnwidth}
		\centering
		\includegraphics[width=\linewidth]{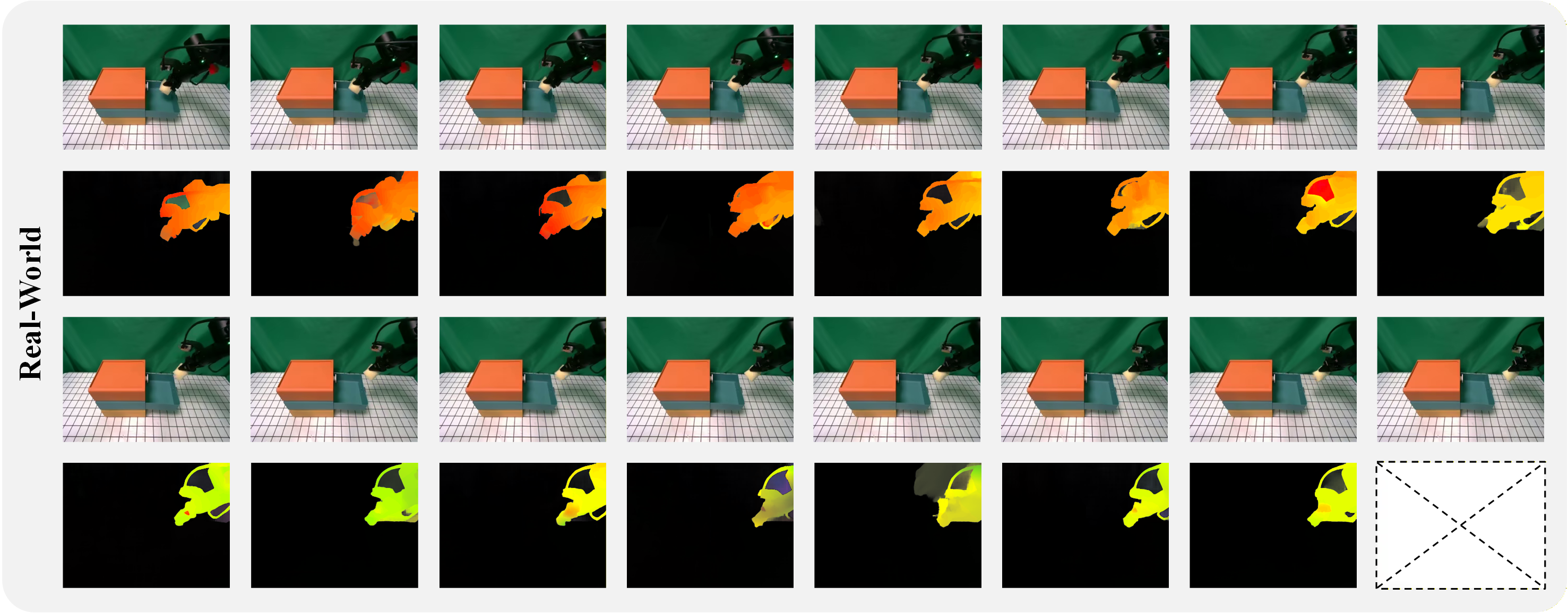}
	\end{minipage}
	\caption{Real-World experiments were conducted on an ARX-R5 robot for a drawer-opening task, which involves opening the drawer, retrieving a sponge, and placing it on the table. Data were collected at 20 Hz over 100 trajectories (47,989 samples) across five different background settings, with each evaluation repeated 20 times.}
	\label{fig:realworldlaof}
\end{figure}

\begin{table}[h!]
	\centering
	\caption{
		Canonical Correlation Analysis between optical flow representations and physical actions across different models. Optical flow features were first extracted using DINOv2 (768-dim) and then reduced to 50 dimensions via Principal Component Analysis (PCA) before computing CCA. Each benchmark reports results over 200K randomly drawn samples. Here, $\Delta$DINOv2 denotes features obtained by differencing consecutive observations extracted using DINOv2. “Noise” denotes Gaussian noise, and “Action” denotes physical actions.
	}
	\renewcommand{\arraystretch}{0.8}
	\setlength{\tabcolsep}{5pt} 
	\resizebox{0.55\linewidth}{!}{%
		\begin{tabular}{lcccccc}
			\toprule
			\multirow{2.5}{*}{\centering Method} 
			& \multicolumn{2}{c}{LIBERO-Plus} 
			& \multicolumn{2}{c}{PROCGEN} 
			& \multicolumn{2}{c}{Real-World} \\
			\cmidrule(lr){2-3} \cmidrule(lr){4-5} \cmidrule(lr){6-7}
			& Action & Noise & Action & Noise & Action & Noise \\
			\midrule
			RAFT & 0.67 & 0.03 & \cellcolor{red!10}\textbf{0.78} & 0.03 & 0.61 & 0.06 \\
			SEA-RAFT & \cellcolor{red!10}\textbf{0.68} & 0.03 & 0.76 & 0.03 & \cellcolor{red!10}\textbf{0.65} & 0.06 \\
			$\Delta$DINOv2 & \cellcolor{blue!10}\textbf{0.43} & 0.03 & \cellcolor{blue!10}\textbf{0.42} & 0.03 & \cellcolor{blue!10}\textbf{0.45} & 0.06 \\
			\bottomrule
	\end{tabular}}
	\label{tab:cca}
\end{table}

\begin{table}[h!]
	\centering
	\caption{Experimental results of different optical flow models on LIBERO-Plus and Real-World. LIBERO-Plus \cite{fei2025libero} extends LIBERO by including robustness evaluations under distractors. The models were trained on mixture data (2M samples) and evaluated under Light conditions (variations in intensity, direction, color, and shadow) and Background conditions (changes in scene and surface appearance). Here, LAOF-R denotes optical flow extracted using RAFT, LAOF-S denotes SEA-RAFT, and LAOF-$\Delta$ denotes features obtained by differencing consecutive observations extracted using DINOv2
	}
	\renewcommand{\arraystretch}{0.8}
	\setlength{\tabcolsep}{5pt} 
	\resizebox{0.55\textwidth}{!}{
		\begin{tabular}{lcccc}
			\toprule
			Method & LIBERO & LIBERO-Plus & PROCGEN & Real-World \\
			\midrule
			LAPO      
			& \cellcolor{blue!10}$\mathbf{72.6}$ 
			& \cellcolor{blue!10}$\mathbf{43.4}$
			& \cellcolor{blue!10}$\mathbf{0.535}$ 
			& \cellcolor{blue!10}$\mathbf{14/20}$ \\
			
			LAOF-R      
			& $76.8$
			& ${63.5}$
			& \cellcolor{red!10}$\mathbf{0.693}$
			& \cellcolor{red!10}$\mathbf{18/20}$ \\
			
			LAOF-S  
			& \cellcolor{red!10}$\mathbf{78.4}$ 
			& \cellcolor{red!10}$\mathbf{68.3}$
			& ${0.651}$
			& \cellcolor{red!10}$\mathbf{18/20}$ \\
			
			LAOF-$\Delta$ 
			& $73.7$ 
			& $57.7$
			& $0.544$
			& $16
			/20$ \\
			\bottomrule
		\end{tabular}
	}
	\label{tab:method_comparison}
\end{table}

\subsection{LIBERO-Plus and Real-World Experiments}
As shown in Table \ref{tab:method_comparison}, LAOF consistently improves robustness over LAPO, regardless of the specific optical flow model employed. This improvement holds across different environments and evaluation settings, indicating that incorporating optical flow constraints into latent action learning provides a systematic advantage under distribution shifts and visual perturbations. A comparison with Table \ref{tab:cca} further reveals a clear relationship between motion–action alignment and downstream performance. On PROCGEN, RAFT achieves higher Canonical Correlation Analysis (CCA) correlations between optical flow representations and physical actions, which coincide with superior task performance. In contrast, on LIBERO-Plus, SEA-RAFT attains higher CCA correlations and correspondingly better overall results. Notably, even LAOF-$\Delta$, which replaces explicit optical flow estimation with simple differencing of consecutive DINOv2 features, achieves competitive performance. These results indicate that when motion representations more effectively capture action-relevant dynamics, the resulting latent action representations become more informative and inherently more robust to visual distractors.

\subsection{Ablation Study on $\lambda$}
\begin{figure}[h!]
	\centering
	\begin{subfigure}{0.24\textwidth}
	\centering
	\def\svgwidth{0.99\linewidth}
	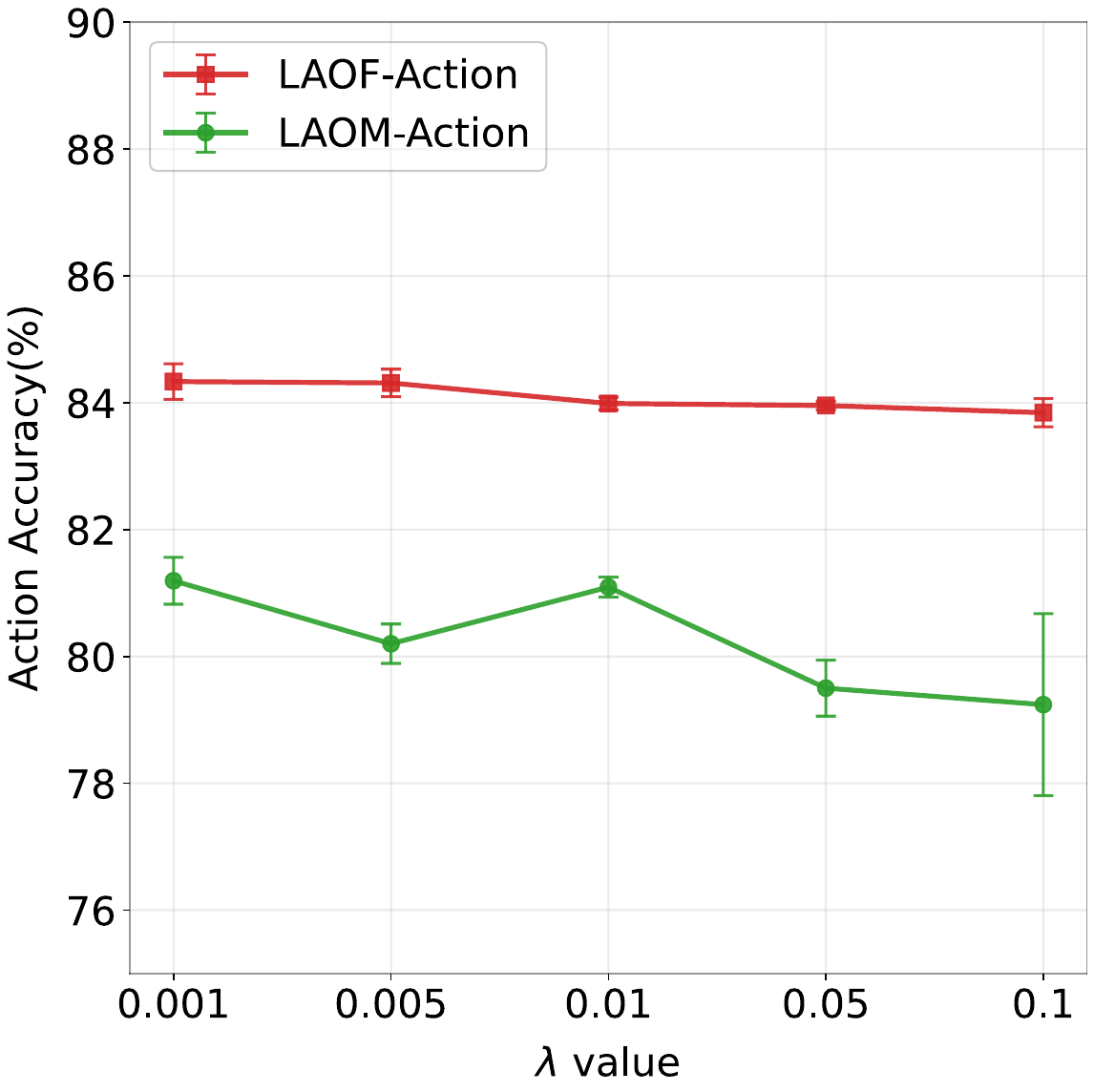
	\caption{BIGFISH}
	\label{fig:lambda_bigfish}
	\end{subfigure}
	\hfill
	\begin{subfigure}{0.24\textwidth}
		\centering
		\def\svgwidth{0.99\linewidth}
		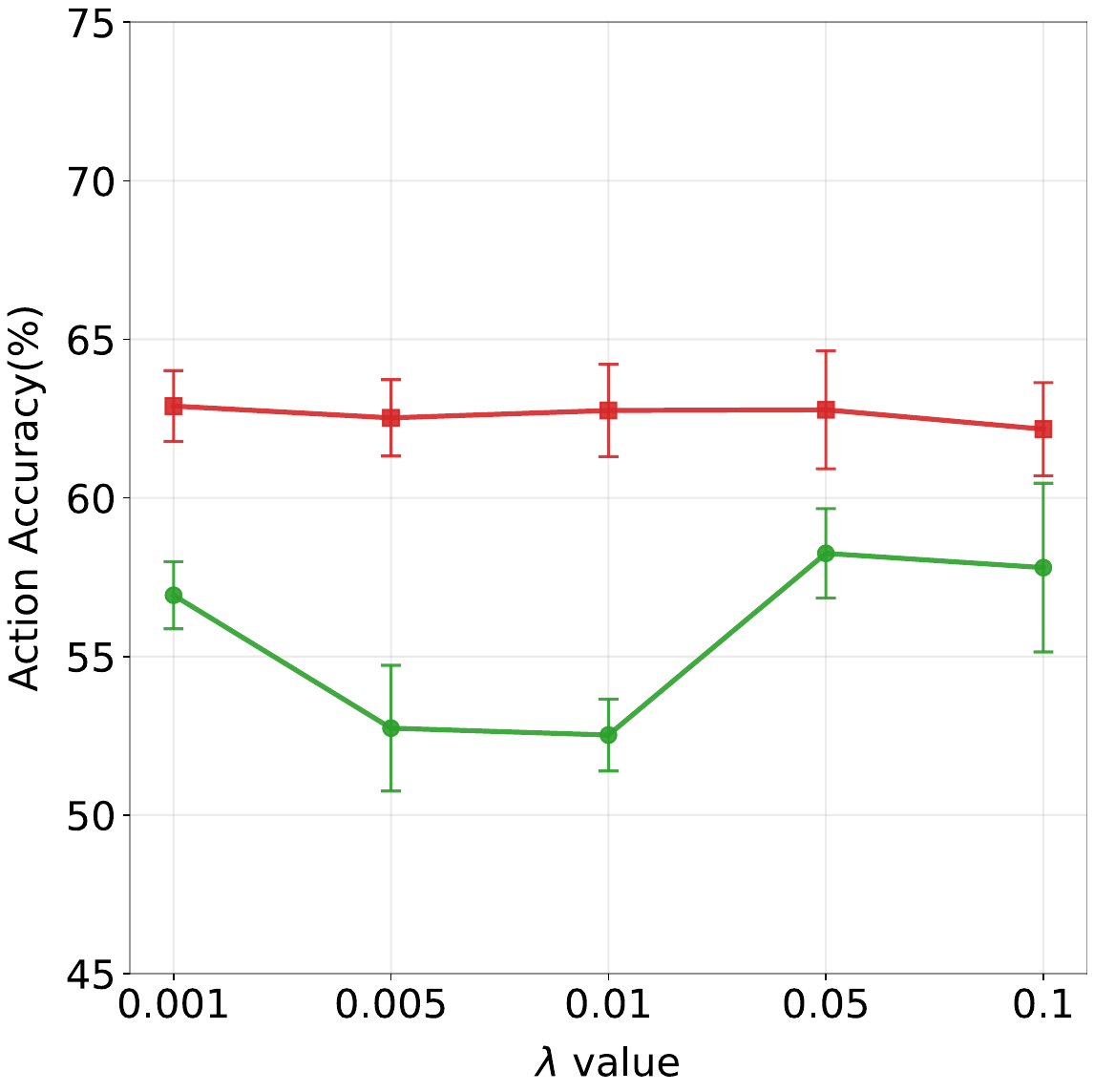
		\caption{CHASER}
		\label{fig:lambda_chaser}
	\end{subfigure}
	\hfill
	\begin{subfigure}{0.24\textwidth}
	\centering
	\def\svgwidth{0.99\linewidth}
	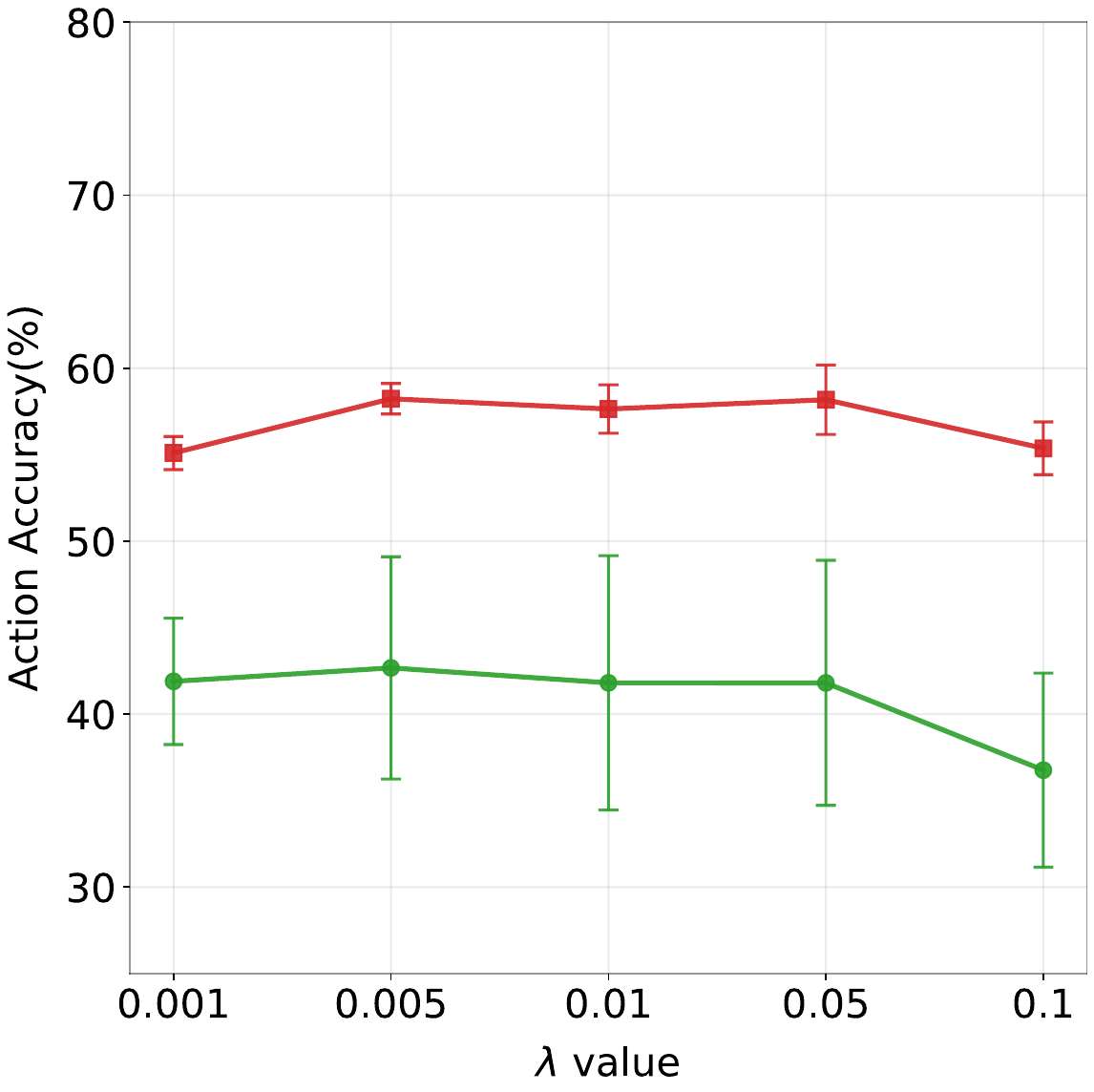
	\caption{LEAPER}
	\label{fig:lambda_leaper}
	\end{subfigure}
	\hfill
	\begin{subfigure}{0.24\textwidth}
	\centering
	\def\svgwidth{0.99\linewidth}
	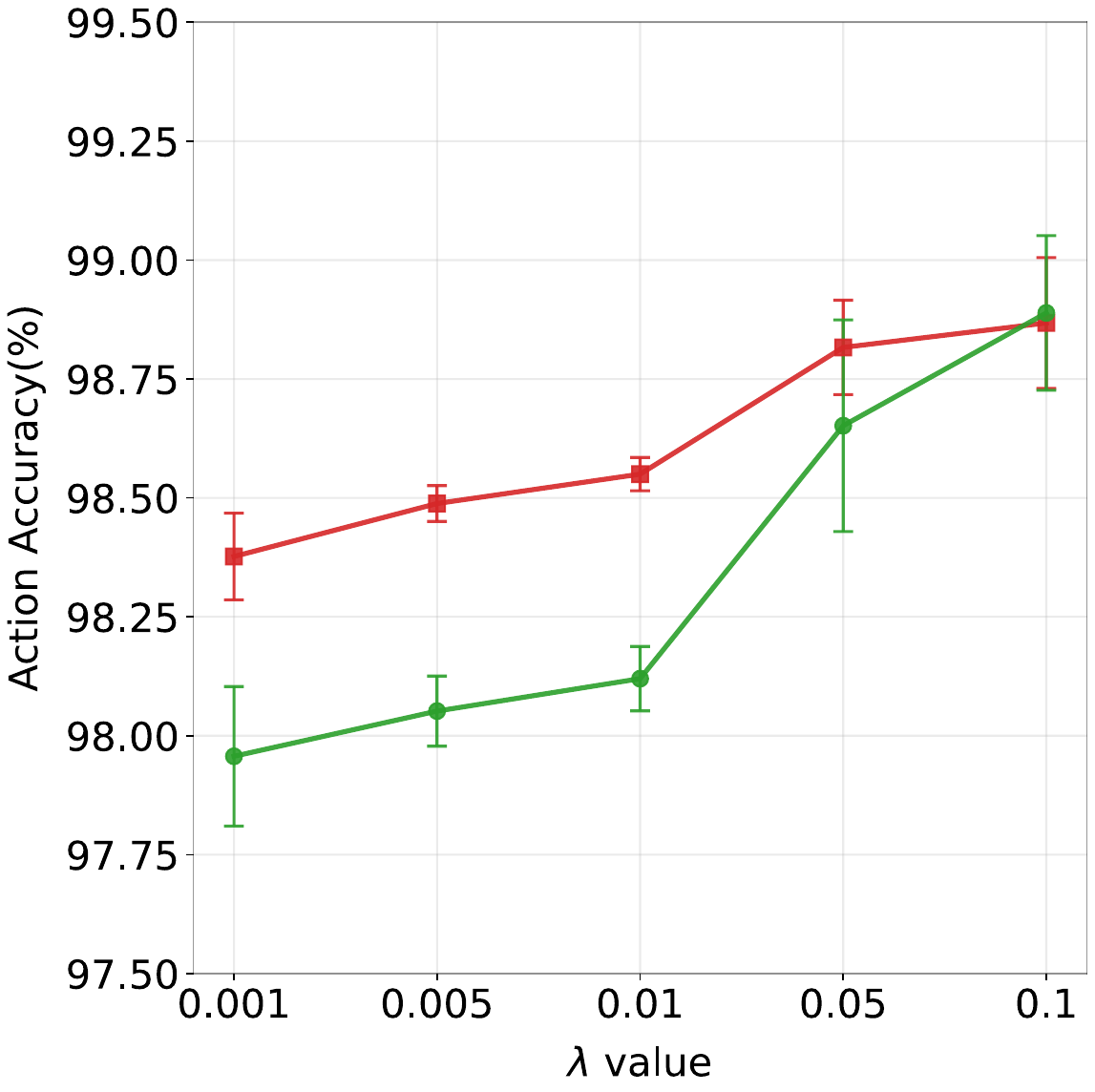
	\caption{HEIST}
	\label{fig:lambda_heist}
	\end{subfigure}
	
	\caption{Effect of the coefficient $\lambda$ on training stability. A shared legend for all four subfigures is shown in subfigure (a).
	}
	\label{fig:lambda_all}
\end{figure}

The coefficient $\lambda$ balances the contributions of action supervision and optical flow constraints. In our experiments, we adopt the default setting $\lambda = \frac{M}{N + M}$, reflecting the proportion of action-labeled data available during training. To eliminate the confounding influence of $\lambda$ on assessing training stability, we conduct an ablation study under a fixed action ratio of 1$\%$, varying the coefficient across roughly an order of magnitude, i.e., [0.001, 0.1]. We report the mean and standard deviation of action accuracy in Fig.~\ref{fig:lambda_all} to assess its influence on the learning process. The results indicate that LAOF-Action remains highly stable across the entire range, exhibiting consistently lower variance and smoother performance changes. In contrast, LAOM-Action demonstrates substantially higher sensitivity, with notable performance fluctuations and clear signs of training instability. These findings show that LAOF-Action is significantly more robust to changes in $\lambda$ and provides more reliable learning behavior under sparse action supervision, eliminating the need to learn $\lambda$. Future work could explore making $\lambda$ a learnable parameter to potentially achieve optimal performance.

\begin{table}[h!]
	\centering
	\caption{Hyperparameters, dataset sizes, and learning rates for tasks (action ratio = 1$\%$)}
	\begin{tabular}{lcccccl}
		\toprule
		Task & $\sigma$ & $\lambda$ & Training Samples & Testing Samples & LangSAM Prompts & Learning Rate \\ 
		\midrule
		SPATIAL & 0.05 & 0.01  & 47,056 & 6,173 & - & 
		\multirow{4}{*}{\shortstack[l]{Stage1: 1e-4 \\ Stage2: 3.5e-4 \\ Stage3: 3.5e-4}} \\ 
		OBJECT  & 0.05 & 0.01  & 59,860 & 7,449 & - &  \\ 
		GOAL    & 0.05 & 0.01  & 46,444 & 6,451 & - &  \\ 
		LONG    & 0.05 & 0.01  & 90,695 & 13,585 & - &  \\ 
		\midrule
		BIGFISH & 0.01 & 0.01  & 574,566 & 58,073 & green & \multirow{4}{*}{\shortstack[l]{Stage1: 3e-4 \\ Stage2: 2e-4 \\ Stage3: 3e-5}} \\
		CHASER  & 0.01 & 0.01  & 820,762 & 68,605 & purple &  \\
		LEAPER  & 0.005 & 0.01 & 356,685 & 39,063 & orange-white hair &  \\
		HEIST   & 0.01 & 0.01  & 342,590 & 36,741 & green-frog &  \\
		\bottomrule
	\end{tabular}
	\label{tasks_hyperparameters}
\end{table}

\section{Implementation Details}
\subsection{Task Setup}
For each task, we adopt task-specific environment hyperparameters, as detailed in Table \ref{tasks_hyperparameters}. Instead of using the full resolution normalization term $\sqrt{H^2 + W^2}$, we introduce a coefficient $\sigma$ to avoid excessively attenuating subtle motions and to ensure sensitivity to small but meaningful movements. Since the magnitude of agent motions controlled by ground-truth actions differs across environments, $\sigma$ is set accordingly. For LIBERO, the robotic arm control magnitudes are consistent across tasks, so the same $\sigma$ is used. In contrast, for PROCGEN tasks such as LEAPER, the control dynamics differ from other games. The dataset is divided into separate training and testing splits. The model is trained solely on the training set, while the testing set is used exclusively for evaluation, where we report either the accuracy or the mean squared error of decoding latent actions back into physical actions. Furthermore, to obtain high-quality optical flow, the initial data are collected at a resolution of 512×512, and the corresponding optical flow is extracted using RAFT \cite{teed2020raft}. For data from the PROCGEN benchmark, additional processing with LangSAM \cite{langSAM} is applied to obtain object-centric optical flow. Finally, for LIBERO and PROCGEN datasets, both the original images and the generated optical flow are resized to 256×256 for storage.

\subsection{Model Setup}
We use a frozen DINOv2 \cite{oquab2023dinov2} encoder (ViT-B/14) to extract images features from normalized input frames. Task instructions are encoded using a frozen T5 \cite{raffel2020exploring} text encoder, which provides language embeddings that condition the downstream model components. The action decoder is implemented as a lightweight multilayer perceptron. For LIBERO tasks, the IDM is implemented as a spatio-temporal transformer \cite{xu2020spatial}, while both the FDM and the flow decoder adopt spatial transformers. For PROCGEN tasks, since we do not employ a multi-task model and each task must be trained independently, we use a simpler architecture to accelerate the extensive ablation studies: the IDM is implemented as a lightweight CNN encoder, and both the FDM and the flow decoder are directly implemented using a U-Net \cite{ronneberger2015u} architecture. To investigate different forms of latent action representations, we use a VQ-VAE \cite{van2017neural} for the discrete setting, and a standard VAE for the continuous setting, with the KL-divergence term weighted by 1e-6.

\end{document}